\documentclass[a4paper,11pt]{scrartcl}
\usepackage{geometry}
\usepackage{natbib}
\newcommand{\comment}[1]{\text{\quad\quad$\rhd$ #1}}
\usepackage{booktabs}
\usepackage[noend]{algpseudocode}
\usepackage[utf8]{inputenc} 
\usepackage[T1]{fontenc}
\usepackage[dvipsnames]{xcolor}
\usepackage{graphicx}
\usepackage{j_notation}
\usepackage{my_algorithm}
\usepackage{j_writing}
\usepackage{cancel}
\usepackage{url}
\usepackage{hyperref}
\newcommand{\basepath}{/home/jesse/MEGA/Travail/Notes/}
\renewcommand{\basepath}{.}
\usepackage{subfigure}
\graphicspath{{\basepath}}
\usepackage{hyperref}

\begin{document}

\title{
	From Multi-label Learning to Cross-Domain Transfer: A Model-Agnostic Approach
}  

\author{Jesse Read\\\small\texttt{jesse.read@polytechnique.edu}}
\date{\small LIX Laboratory, Ecole Polytechnique, Institut Polytechnique de Paris, France}
\maketitle

\abstract{
	In multi-label learning, a particular case of multi-task learning where a single data point is associated with multiple target labels, it was widely assumed in the literature that, to obtain best accuracy, the dependence among the labels should be explicitly modeled. This premise led to a proliferation of methods offering techniques to learn and predict labels together, for example where the prediction for one label influences predictions for other labels. 
	Even though it is now acknowledged that in many contexts a model of dependence is not required for optimal performance, such models continue to outperform independent models in some of those very contexts, suggesting alternative explanations for their performance beyond label dependence, which the literature is only recently beginning to unravel. 
Leveraging and extending recent discoveries, we turn the original premise of multi-label learning on its head, and approach the problem of joint-modeling specifically under the absence of any measurable dependence among task labels; for example, when task labels come from separate problem domains. We shift insights from this study towards building an approach for transfer learning that challenges the long-held assumption that transferability of tasks comes from measurements of similarity between the source and target domains or models. This allows us to design and test a method for transfer learning, which is model driven rather than purely data driven, and furthermore it is black box and model-agnostic (any base model class can be considered). We show that essentially we can create task-dependence based on source-model capacity. The results we obtain have important implications and provide clear directions for future work, both in the areas of multi-label and transfer learning.  
}

\section{Introduction}
\label{sec:intro}

Tackling multiple tasks in a single machine learning framework is becoming commonplace, and is found in different forms, and applied to a wide variety of problems. Specific cases of multi-task learning include multi-label classification, where tasks are binary classification problems, sharing common input; and transfer learning, where only one or a subset of tasks will be evaluated and the other tasks are leveraged to obtain better performance (predictive performance, computational performance, or both) under that evaluation.


The \textit{raison d'être} of these areas of the literature stems from the same promise: by formulating tasks together, we can obtain greater predictive performance and/or reduced computational requirements, than if we approach each task separately. 
To a significant degree, the promise has been met. The multi-label literature, for example, has flourished over the previous decades with a plethora of successful learning methods that out-compete independent models. 

However, the success of considering multiple tasks jointly rather than separately has been less widespread in, for example, transfer learning (even if the success has been just as significant; it is nevertheless less widely employed; more restricted to particular domains than, say, multi-label learning) due to a number of major challenges invoked by one assumption: that the tasks in consideration should exhibit similarity. In multi-label learning, since tasks share a common input and a common dataset, there is usually some assumed inter-dependence among the output labels. In transfer learning, the source task does not form an integral part of the target task, rather one needs to identify a source task of adequate similarity. This increases manual and computational search time (the savings of which was a primary motivator for considering transfer learning in the first place). 

Purely data-driven approaches can be costly, and for than reason model-based transfer-learning is an interesting alternative. Recent developments have almost invariably taken the form of novel deep neural architectures, and have met great success in domains such as natural language and images, where data is large, and deep learning models are notoriously successful yet also too expensive to train from scratch on all problems due to the enormous quantities of data needed to be processed. Therefore the current common practice is essentially importing and (optionally) fine-tuning hidden layers from other neural networks and this practice is for a restricted small set of domains. 

However, very many machine learning engineers and data scientists around the world use algorithms that do not fall into the category of `deep learning' (\cite{AnEstimate} estimates 600,000 monthly scikit-learn users -- a framework which contains only rudimentary neural network learning algorithms). Yet, many machine learning models cannot be easily dissected into hidden layers expressed as weight matrices as in popular deep-learning frameworks, and even when such an expression is trivial, sharing such layers is not yet widely practiced (at least; not beyond the image and natural-language domains). 




In this article we approach the problem of model-based model-agnostic transfer learning (i.e., model-driven/no source data available, and no possibility to inspect the models). The key component of our approach is to remove the assumption of task similarity, i.e., the source and target tasks may come from \emph{unrelated} domains. More precisely: we do not require any more similarity between target and source domains, than between any randomly selected domains in the scope of any task having somewhere some useful application.

We do this by building from aspects and observations that have arisen from the multi-label literature. Notably, in this domain, model-agnostic approaches are commonplace (where they are often known as \emph{problem transformation} or data transformation methods); implicating a wide diversity of model classes (decision trees, support vector machines, logistic regression, etc.). These models are often adapted to produce predictions for multiple tasks simultaneously, or separate instantiations of such models are linked together across the tasks via their predictions in a modular fashion within a `meta' method. 
Such methods are widely reported to outperform independent models. Yet, interestingly, one can observe that these empirical studies indicate such outperformance (of interconnected dependence-based models vis-a-vis independent models) even when label dependence is not explicitly required to minimize a loss metric. This allows us to formulate a new hypothesis: transfer learning can still be advantageous even when target tasks come from unrelated application domains. 

This is an extremely challenging context. Because the model representation is out of reach we can only access inputs and outputs and thus face an acute information bottleneck. We overcome this challenge by generalizing to multi-label transfer learning, where source tasks are multi-label tasks, producing an output vector of predictions, of sufficient size and richness to convey some additional capacity to the target model. Since the source data is unavailable, to obtain these predictions we must feed the source model with \emph{target} inputs. This embodies the main challenge: source and target domains are unrelated to each other, so we must use the source task as an analogy to the target task; creating a kind of artificial dependence between the tasks and making use of similarity that occurs by chance. 

In other words this paper looks to translate some important results from multi-label learning into the particular case of model-agnostic cross-domain transfer learning; and then study and develop them further in this context. The main objective is not to forward a number of empirical outcomes on benchmark data sets supporting a claim to a new state of the art method, but rather to study and demonstrate underlying mechanisms, and discuss their implications. We develop a deeper understanding on how task similarity/correlation is involved in multi-task learning, from the perspective of multi-label classification. Alongside empirical testing, we discuss the implications on different types of transfer learning, including, for example, adaptation to concept shift in data-streams. 


The article is outlined as follows: In \Sec{sec:problem_setup} we provide notational preliminaries and formally define the problem setting. In \Sec{sec:transfer} we review transfer learning and discuss particular methodologies of interest, looking also at related areas relevant to our study such as multi-task and multi-label learning, and concept drift adaptation. In \Sec{sec:multilabel} we draw lessons from the existing (and relatively well-developed) area of multi-label learning, and use them to formulate a number of hypotheses towards our goal of model-agnostic cross-domain transfer learning. In \Sec{sec:development} we take these lessons and apply them to develop a novel approach to transfer learning under the numerous constraints and motivations outlined above; named Transfer Chains. Both sections \Sec{sec:multilabel} and \Sec{sec:development} are supplemented by a number of empirical evaluations within. We pool together outcomes and discuss at length in \Sec{sec:discussion}, leading to number of observations and recommendations that have important implications in the general area of transfer learning.  
We conclude in \Sec{sec:conclusion}, outlining some promising future directions. 



\section{Notation and Problem Setting}
\label{sec:problem_setup}


A task is defined as seeking to build model 
\[
	h : \dX \rightarrow \dY
\]
mapping instances $\x$ from an input domain $\dX = \R^d$ to labels $\y$ belonging to the output domain $\dY = \R^m$, under the goal of minimizing some loss metric $\ell : \dY \times \dY \rightarrow \R$. In the multi-label case, $m > 1$ (typically, in the multi-label classification setting, $\dY = \{0,1\}^m$). In the multi-task setting, we can explicitly denote a number of tasks $1,2,\ldots$, i.e., $\dX_1,\dX_2,\ldots$ and $\dY_1,\dY_2,\ldots$. Note that there is also ambiguity here: the multi-label setting could  be defined as a kind of multi-task setting having a single $\dX$ and multiple task labels $\dY_1,\dY_2,\ldots$; one for each task/label. And, even, $\dY_1 = \dY_{1,1} \times \ldots \times \dY_{1,m_1}$ (multi-dimensional labels). We embrace this ambiguity throughout, in order to bridge between multi-label and multi-task settings (which are otherwise often treated separately in the literature); the distinction is only in the input: a multi-label task has a single domain $\dX$. Transfer learning is the specific multi-task task where a number of tasks are source models and the others are target tasks. For notational simplicity we denote a single source task ($S$) and single target task ($T$), but without loss of generality; indeed in experiments we consider several source tasks.

We approach the task of model-based model-agnostic transfer, implying the construction of target model $h_T$, leveraging a source model $h_S$, under the unique combination of constraints that we:  
\begin{enumerate}
	\item do not have access to the source data set(s), 
	\item cannot inspect the source model(s); and 
	\item the source domain is not related to the target domain. 
\end{enumerate}
Note that this is different from, and much more difficult than, the huge majority of transfer learning approaches which seek to find and inspect source tasks, measure their similarity to the target task, then transfer relevant parts of either into the source model, prior to (possible) fine tuning. 

We are supplied with 
\begin{enumerate}
	\item A source model $h_S : \dX_S \rightarrow \dY_S$ (or, generally, models) already built to tackle some source task
	\item A target data set $\D_T$ 
\end{enumerate}
and the goal is to build a model for the target task  
\(
	h_T : \dX_T \rightarrow \dY_T.
\)

Despite the unique combination of constraints our general objective is to nevertheless the same; to make use of source models in the form of transfer learning to improve accuracy on the target task. 


There is no strict requirement to use a source model $h_S$ at all when constructing the target model $h_T$, yet we are motivated to do so by the potential to improve learning (as opposed to a purely data-driven approach to the target task), including \cite{TransferLearningBook} that predictive performance: 
\begin{enumerate}
	\item is initially higher (after initialization) 
	\item increases more quickly (during training); and 
	\item plateaus at higher level (when training is finished).
\end{enumerate}
We investigate if such an advantage can hold after removing the common assumption that source task and target task are in some way related. This is the far less-common \textit{cross-domain} transfer learning. 
We assume this setting to the extent that we do not expect any more similarity between the target task and source task than would be present between any two tasks drawn uniformly at random from all possible real-world datasets. These strict assumptions make it infeasible to compete with modern state of the art transfer-learning methods (which explicitly assume and aim to leverage similarity). But they raise interesting considerations of if/when, and how, it can be beneficial to carry out transfer learning in this setting. We investigate these considerations.  


\section{Related Work}
\label{sec:transfer}
\label{sec:related}

\begin{figure}
	\centering
	\subfigure[Multi-Label]{%
		\includegraphics[scale=1]{\basepath/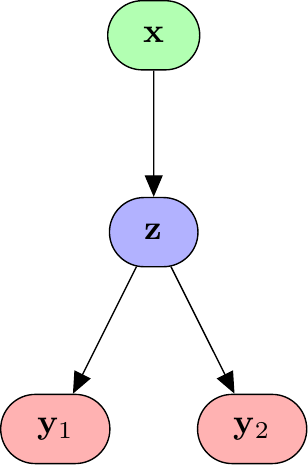}  
		\label{fig:deep_br}
	}\quad
	\subfigure[Multi-Task]{%
		\includegraphics[scale=1]{\basepath/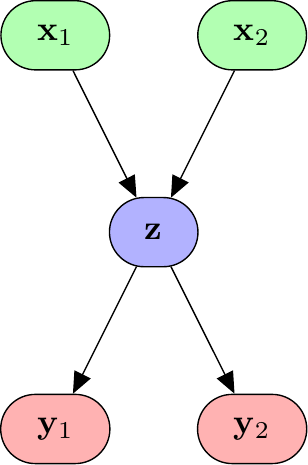}  
		\label{fig:mt_nn}
	}\quad
	\subfigure[Transfer]{%
		\includegraphics[scale=1]{\basepath/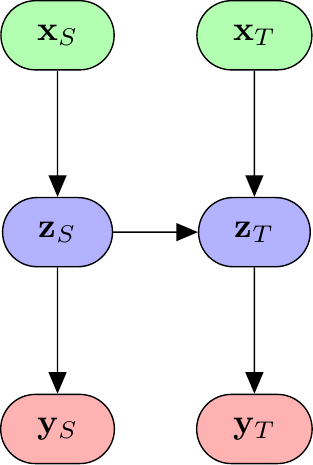}  
		\label{fig:tl}
	}\quad
	\caption{\label{fig:intro}Simplified architectures (with hidden representation, $\z$) for \fref{fig:deep_br} multi-label learning, \fref{fig:mt_nn} multi-task learning, and \fref{fig:tl} transfer learning. 
	In \fref{fig:tl} we see two intermediate $\z$-layers, but only one is used per task. In fact, the multi-label task is a generalization, having multi-dimensional labels; $\y_1 \in \R^{m_2}$; $\y_2 \in \R^{m_2}$).}
\end{figure}

The task we approach is essentially a heavily constrained type of transfer learning. In this section we look at links to existing work on transfer learning in the literature as relevant to this setting. Table~\ref{tab:1} places model-based transfer learning in the context of related tasks. Recall, we are specifically interested in the case where knowledge is stored in the form of existing models (rather than data sets). Data-driven approaches are enormously important and have advanced the applicability of machine learning in many areas, but do not coincide with our particular constraints. Meanwhile, \Fig{fig:intro} exemplifies the differences (and similarities) among different models.  


\begin{table}[h!]
\centering
	\caption{\label{tab:1}Transfer learning and related methodologies; denoted with respect to two potential tasks: 1 and 2, when the tasks may be learned concurrently and are both of interest, and tasks $S$ (source) and $T$ (target) when the target task is the main focus. In self-taught learning, only data from input-domain of the source task is available.  In transfer learning, input data and/or source models can be available; in our case we consider only source models and no data.}
\begin{tabular}{llll}
	\hline
		Name                 & Target Task                           & Source Task                                                                                                                                                                        &  Comment \\
	\hline
	[Semi-]Supervised learning      & $\dX \mapsto \dY$                          &                                                                                                                                                                                       & Single task \\
	Multi-label learning     & $\dX \mapsto \{\dY_1,\dY_2\}$                & & Several tasks, $\dX$ in common \\ 
	Multi-task learning      & $\{\dX_1,\dX_2\} \mapsto \{\dY_1,\dY_2\}$ &  & Several tasks \\ 
	Transfer learning        & $\dX_T \mapsto \dY_{T}$                      & $\dX_S \mapsto \dY_{S}$                                                                                                                                                                   &  Source and target tasks \\
	Concept-drift (input-only)         & $\dX_S \rightarrow \dY_{T}$                & $\dX_S \mapsto \dY_{S}$                                                                                                                                                                 &  $\dX$ in common, models available \\
	Concept-drift (output-only)         & $\dX_{T} \rightarrow \dY_S$                & $\dX_S \mapsto \dY_{S}$                                                                                                                                                                 & $\dY$ in common, models available  \\
	Self-taught learning     & $\dX_T \mapsto \dY_T$                          & $\dX_S {\color{gray} \mapsto \dY_S}$                                                                                                                                                                                 & Source task unknown \\ 
	\hline
	\end{tabular}
\end{table}

\subsection{Types of transfer learning and methods}
\label{sec:types}

The canonical task of transfer learning in machine learning, from which many diverse problem settings arise, can be phrased as transferring knowledge from a \emph{source task} to a \emph{target task}. The transfer can be in the form of data (data-based transfer), a model representation (model-based transfer), or even hyper-parameters (often called in this form meta learning). A general survey and introduction is given by \cite{TransferLearning,TransferLearningBook}.

Deep transfer learning (DTL) (a survey by \cite{DeepTransferLearningSurvey}) is by far the most common form of transfer today, owing to the enormous current popularity and renown effectiveness of deep learning. It is a straightforward, well-documented, and largely successful recipe for transfer learning, usually cast as follows (and depicted in \Fig{fig:tl}): 
\begin{enumerate}
	\item Identify a suitable deep neural network source model $h_S : \dX_S \rightarrow \dY_S$; where source domain $\dX_S \times \dY_S$ is related in some way to the target domain $\dX_T \times \dY_T$; 
	\item  transplant one or several hidden layers (depicted as $\z_S \mapsto \z_T$ in \Fig{fig:tl}) from the source architecture into the target architecture; and 
	\item train ($\z_T \mapsto \y_T$) and -- optionally -- fine tune the full target network ($\x_T \mapsto \z_T \mapsto \y_T$). 
\end{enumerate}

We can cite the image vectors of \cite{VGG} and word-embeddings of \cite{word2vec} as approaches having an enormous impact on the community, especially in the areas of computer vision and natural language processing where the representations of these models (and others of similar nature) are often incorporated into new models as a standard, or even necessary component for producing state of the art models. Recent work incorporating deep-learning frameworks such as attention mechanisms \citep{TransferLearningAttention} has further improved this area. 

In the probabilistic sense, transfer learning can be carried out by using the posterior distribution of the source model as a prior for another. By Bayes' theorem\footnote{Throughout, we use the notation $P(Y|x)$ as shorthand for $P(Y|X=x)$; the conditional distribution -- conditioned on observation $x$; a realization of the random variable $X$}, 
\[
	P(\Y_T|\x_T) = \frac{P(\x_T|\Y_T) P(\Y_T)}{P(\x_T)} \propto P(\x_T|\Y_T) P(\Y_T)
\]
where we then use $P(\Y_T) \approx P(\Y_S|\x_T)$ from the posterior of the source task. 
And more specifically, in Bayesian methods we may place a prior over model parameters \citep{OptimalBayesTransfer,BayesianNeuralTransfer}. 

Concept drift and -- particularly in this context -- the \emph{adaptation to} concept drift \citep{IndresSurvey,DriftModelReuse} is a central concept to a large volume of literature on learning from data streams. Even if not typically discussed explicitly by authors, adaptation to drift is indeed a form of transfer learning.  Namely, the target task (i.e., target \emph{concept}) is a degradation of the source task wrt that task. 
This form of transfer learning is interesting to us, because sudden and complete concept shift (considered frequently in the literature \citep{IndresSurvey}), meaning that source and target domains are unrelated, is equivalent to the scenario we are aiming to tackle.   
The majority of the data-stream literature suggests the best course of action is to scrap completely the earlier (source) models. Our results specifically challenge this view (discussion later in \Sec{sec:insights_streams}).

This area of the literature is also interesting to our objectives because methods for drift-detection and adaptation are often treated as black-box and model-agnostic; typically a dynamic ensemble of models is maintained, without the need to explicitly specify the base model class \cite{DataStreamEnsembleSurvey} and members are reset (simply purged and reinitialized) over time. 
Maintaining models in a data stream means that, almost by definition (since streams are potentially infinite, whereas resources are usually not), source data cannot be kept for the source task; thus driving model-driven adaptation. 

The assumption of that source data is not available has also been tackled in, e.g., the context of natural language processing by \citep{ModelBasedNLPTransfer}, an approach both multi-source and model-based (i.e., not requiring parallel data). And \cite{DeepTransferMarkovLogic} consider the case of different distribution and different domain, finding structural similarities in the form of Markov logic. Meanwhile, \cite{CrossDomainAdaptation} perform cross-domain adaptation by focusing not only on a transfer of features, but rather of \emph{metrics} of similarity. The importance of task diversity (as opposed to the importance of task similarity which is usually of focus) is approached by \cite{TheoryTLTaskDiversity}, but they still rely on the common assumption that the main purpose is to learn a feature representation shared across different tasks  (i.e., multi-task, rather than transfer learning). 



Meta-learning is a kind of transfer learning that transfers from the \emph{meta} source task in terms of hyper-parametrization, including particular architecture search. In other words, knowledge from the source domain itself is not directly transferred, but rather, knowledge of how to approach tasks.  There are a number of modern examples (e.g., \citep{MML,MAML,MMAML,MMLshrink,NASRL}). However, despite often carrying the `model-agnostic' label, these methods are still gradient-descent/neural-network centric and almost invariably assume access to source datasets (i.e., they are predominantly data-driven, rather than model-driven). 


\subsection{Modular learning and random projection methods}

The approach we take, with no assumptions on task similarity, brings us close to that of modular random projections. 


Random projections and untrained layers have been used in machine learning for decades, including textbook examples of polynomial regression \citep{EOSL}, random-basis function neural networks \citep{DuSwamy}, the eccentrically-named `extreme learning machines' \citep{ELM} and random kitchen sinks \citep{RandomKitchenSinks}, and -- if we involve recurrent layers -- reservoir computing \citep{ReservoirComputing}. Such techniques have been frequently criticized or, worse, largely ignored by the deep-learning community where gradient-descent and back-propagation are the dominant techniques yet, nevertheless, these other diverse methods continue to persist and are successful in many tasks. 


In our work the source model is not a random projection in the traditional sense that the model parameters are drawn randomly, rather in the sense of the model being trained on a source task which was drawn randomly (rather than guided by a similarity metric). Nevertheless, we face potentially the same challenge of information bottlenecks: even if a large amount of data is moved easily in memory the amount of information (relevant to solving the target task) may be poor. 


The model-agnostic setting of not being able to inspect source models or view the data they were trained on (the context we consider) lends itself naturally to a comparison to modular learning; building a web of structure across black-box models. Such methods are well known in the multi-label literature (e.g., by \cite{StructureSearch,CCReview}, and references therein), where modules are off-the-shelf classifier or regression models (a frozen neural network layer would be a particular case, as by \cite{FreezingLayersNN}) also known as parameter-isolation. Indeed,  
this purely model agnostic and modular view comprises the launching pad for our study, from which we proceed in the following section. 



\section{Lessons from Multi-label Learning and the Case of Chaining Methods}
\label{sec:multilabel}
\label{sec:lessons}


Multi-label learning is a specific case of multi-task learning, and often can be approached as a specific case of transfer learning. Particularly, the case of chaining models. 
The multi-label literature has generally overlooked the potentially close connection to transfer learning. In this section we specifically study the area of overlap between these areas and extract insights to 
apply to the special case of model-agnostic transfer learning without task similarity, that we have detailed above. 


In a multi-label task we learn from data $\{(\x^{(i)},\y^{(i)})\}_{i=1}^n$ where $\x^{(i)} \in \dX = \R^d$ and $\y^{(i)} \in \dY = \R^m$ (including the most common case where $\dY = \{0,1\}^m$; $m$ is the number of labels), with which we want build some model $h$ to provide output/prediction,  
\(
	\ypred = h(\x)
\)
for test example $\x$, where target labels $\ypred = [\yp_1,\ldots,\yp_m]$ refer to a unique $\x$-in common. 
In various places (and in \Fig{fig:the_two}) we generalize to multi-dimensional or meta-labels $\y_1,\y_2$; which bridges the notational gap to multi-label transfer learning: we can consider a model connectedness among blocks $\y_1$ and $\y_2$ as much as among individual labels $y_{1}$ and $y_{1}$ of a single block, without loss of generality. 


\[
\]
\subsection{Multi-label graph and chaining methods}

\begin{figure}[h!]
	\centering
	\subfigure[Full Cascade]{%
		\includegraphics[scale=1]{\basepath/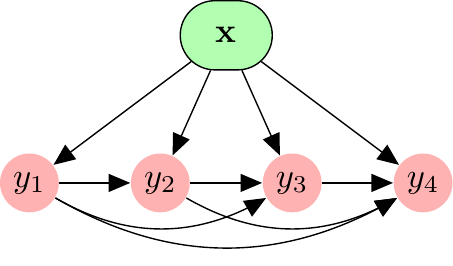}  
		\label{fig:ml_cc_non}
	}
	\subfigure[Modular Chaining]{%
		\includegraphics[scale=1]{\basepath/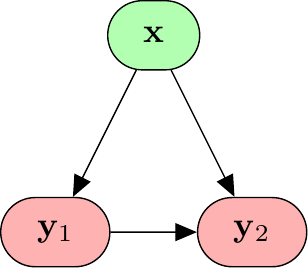}  
		\label{fig:ml_cc}
	}\quad
	\subfigure[Chaining with Deep Architecture]{%
		\includegraphics[scale=1]{\basepath/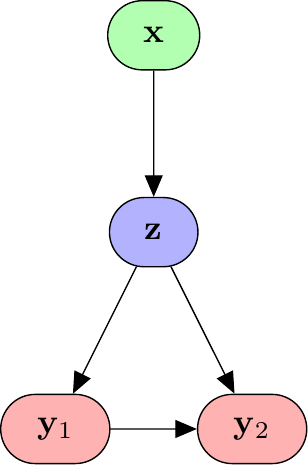}  
		\label{fig:nn_cc}
	}\quad
	\caption{\label{fig:the_two}A depiction of `chaining'; \fref{fig:ml_cc_non} among each of the outputs of a single task, such that $\ypred = [\yp_1,\ldots,\yp_m] = h(\x)$; and also as between two tasks; \fref{fig:ml_cc} with predictions invoked directly from the input, and \fref{fig:nn_cc} via a latent representation. This generalizes to multi-dimensional labels $\ypred_1,\ypred_2 = [\yp_{1,1},\ldots,\yp_{1,m_1},\yp_{2,1},\ldots,\yp_{2,m_2}] = h(\x)$.}
\end{figure}

The fact that (in transfer learning) only the target task is of interest wrt the loss metric, means any construction that the source task need not be considered beyond production of the target model; knowledge is to be transferred only in one direction, from source to target. Many aspects of multi-label learning (which supposes evaluation of all tasks/labels) are clearly not relevant, however, there are indeed classes of models which proceed in a sequential and/or hierarchical fashion, from one completed model to the next, and thus are ideally suited. For example, the family of \textit{chaining methods} (a recent survey in \citep{CCReview}), as deep architectures developed contemporaneously such as ADIOS \citep{ADIOS}, and \citep{LDLS} that follow a chain-like approach. Consider the illustrations in \Fig{fig:the_two}. 

In its simplest rendition of training in chaining for the multi-label learning case, the method first learns   
\begin{align*}
	h_1 &: \dX \rightarrow \dY_1 \text{\quad and then}  \\
	h_2& : \dX \times \dY_1 \rightarrow \dY_2
\end{align*}
in that order (an order which can be inverted in the multi-label case, but not for transfer). Predictions are later obtained from the model as 
\begin{align}
	\ypred_1 &= h_1(\x) \notag \\
	\ypred_2 &= h_2(\ypred_1,\x)\notag  \\
	& = h_2(h_1(\x),\x) \label{eq:x}
\end{align}




We remark that under this chaining mechanism, training of $h_2$ does not require access to $h_1$ or any kind of joint training, because labels $\y_1 \in \dY_1$ are already available in the training data. Further, at test time, $h_2$ only requires access to $h_1$'s predictions (not the data it was trained on, or any knowledge of its gradient or model class). Therefore, this is essentially a model-agnostic transfer from $\y_1$ to $\y_2$. 

We also remark that such an approach is known to perform well (better than independent models) without any strict assumption on dependence/similarity among the label tasks \citep{CCAnalysis,CCReview}. There is only one reason why it does not already meet requirements for transfer learning: the fact that the source $\x \in \dX$ is identical to both $\y_1$ and $\y_2$; whereas we seek performance in the case that $\dX_1 \neq \dX_2$ (inputs from both tasks come from different distributions). Our question for the following is thus: is it possible to extend the chaining approach successfully to the transfer learning scenario without \emph{any} particular assumption of similarity. 


\subsection{The importance of modeling label dependence in multi-label learning}
\label{sec:label_dependence}

Following the seminal paper by \cite{Overview}, the multi-label literature flourished, and over the years that followed many dozens of methods were presented motivated by the goal of modeling labels together 
under the assumption that this was necessary due to the presence of label dependence. This assumption was largely based on intuition and encouraged by empirical results, since indeed, these methods widely improved over the use of independent models. 


This intuition of label relatedness was formalized in the setting of probabilistic dependence by \cite{JointModeEst} (and \cite{OnLabelDependence2}), in particular distinguishing between global label dependence,  
\begin{equation}
	\label{eq:dep}
	P(Y_2|Y_1) \neq P(Y_2)
\end{equation}
(for any two given labels $Y_1$ and $Y_2$); 
and conditional label dependence, after observing an input vector,   
\begin{equation}
	\label{eq:cdep}
	P(Y_2|Y_1,\x) \neq P(Y_2|\x).
\end{equation}
But what was particularly interesting was their analysis wrt loss metrics, which revealed an apparent disconnect between the larger intuitions of the community and its empirical studies, and theoretical results. Namely, it was shown that some common multi-label metrics, such as Hamming loss, could be minimized without any consideration of label dependence, even though dozens of published results (including classifier chains) proclaimed and demonstrated advantages in practice from doing exactly that (as opposed to modeling problems individually). 


Hamming loss (let us denote $\ell_H$) is essentially an average of label-wise losses, thus due to the summation in the average, each label can in theory be tackled individually. On the other hand, the $0/1$-loss (let us denote $\ell_{0/1}$) compares elements such that they must match exactly\footnote{Note that the inverse of Hamming loss and $0/1$ subset loss are known as Hamming score and \emph{exact} match (or subset accuracy), respectively} to obtain $0$ loss, and else a loss of $1$; i.e., a label-vector prediction prediction is treated as a single label; and thus this encourages a model of label dependence, such that accurate combinations can be chosen even at the cost of poorer marginal predictions. The relationship between these two loss metrics is given as: 
\begin{align}
	\ell_H(\y,\ypred) &= \sum_{j=1}^m \indicator{y_m \neq \yp_m} = \ell_{0/1}(y_1,\yp_1) +  \cdots + \ell_{0/1}(y_m,\yp_m)  \label{eq:HL} \\
	\ell_{0/1}(\y,\ypred) &= \indicator{(\sum_{j=1}^m \indicator{y_m \neq \yp_m}) > 0} \label{eq:01} = \indicator{\ell_H(\y,\ypred) > 0}
\end{align}
(where $\indicator{A}=1$ if condition $A$ holds). Many other metrics are possible, but these two are a good illustration on the two extremes, and many methods perform well on both. 

The fact that many methods perform well on both metrics is interesting. Of course, it is true that one metric can act well as a surrogate for another in the multi-label context \citep{dembczynski2010regret,BlendedMetric} (as in the general context of machine learning); i.e., a model that performs well on $0/1$-loss should perform well also on Hamming loss (this is obvious in the extreme case of $\ell_H = \ell_{0/1} = 0$). But, if a classifier has been built to explicitly model label dependence, and performs better than independent classifiers, under Hamming loss which requires no such model, it suggests reasons to model labels together even if they do not demonstrate any form of statistically significant dependence. Indeed in classifier chains which is one such example, it is essentially equivalent to successfully modeling some task labels by taking into other labels whose decision is conditionally independent according to the underlying concept. It is exactly these reasons we wish to study. We do so, alongside novel development, in the following. 


\subsection{Beyond label dependence: other factors behind multi-label performance}

We now ask: if a multi-label architecture that models labels together performs better than models which are fit independently of each other, on a task where labels are known to be independent of each other (both marginally and conditionally wrt $\x$), what explanation is there to this performance? We ask this question so that we might later exploit these mechanisms in a transfer learning setting where source and target tasks are dissimilar. 

\subsubsection{Non-linearity and extra capacity}
\label{sec:non_lin}

For classifier chains in particular it was hypothesised that superior performance (wrt independent models) is partly due to the additional architecture of linking labels together \citep{CCAnalysis}, much like the inner layers of a neural network (where labels in the chain `replace' a latent/hidden-layer representation) \citep{DCC2}. This idea has been recently further developed by \cite{UnifiedView} (more generally for multi-output prediction) and by \cite{CCReview} (more specifically to classifier chain models).

\begin{figure}[h!]
	\centering
	\subfigure[Chain model (a)]{%
	  \includegraphics[scale=1]{\basepath/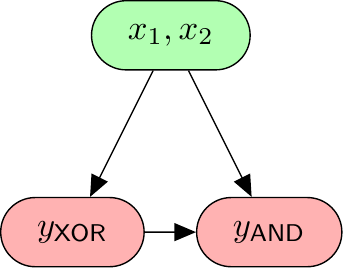}
	  \label{fig:cca}
	  }
	  \quad
	\subfigure[Chain model (b)]{%
	  \includegraphics[scale=1]{\basepath/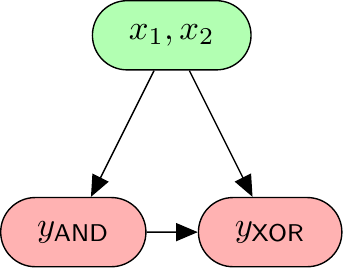}
	  }
	  \quad
	\subfigure[Data from (true, underlying) generating distribution $P$]{%
		\label{tab:xor_1}
	  \begin{tabular}{llll}
		  \hline
			  $X_1$ & $X_2$ &  $Y_{\textsc{xor}}$ & $Y_{\textsc{and}}$  \\
		  \hline                             
			  0     & 0     &  0 &   0    \\ 
			  0     & 1     &  1 &   1    \\ 
			  1     & 0     &  1 &   1    \\ 
			  1     & 1     &  0 &   1    \\ 
		  \hline
	  \end{tabular}
	  }
	  \quad
	\caption{\label{fig:xor_1}A toy example where points of different class labels are  not linearly separable (label \textsc{xor}) and an example where they are (label \textsc{and}). It is not a question of label dependence only, since with a linear base learner (e.g., logistic regression), the \emph{order} of connectivity matters (yet statistical dependence/similarity metrics are symmetric). Rather than a successful model of conditional label dependence, we have gained network capacity from the \textsc{and} node, providing the additional trainable parameter needed to model \textsc{xor}. 
	}
\end{figure}

This can be illustrated by the simplistic problem in \Fig{fig:xor_1} (adapted from \citep{DCC2}; and further developed in the following) and most particularly the model depicted in \Fig{fig:cca}, which is closely related to the classic solution of neural networks for the logical \textsc{xor} function. However, unlike the classic example, this is done in a chain not by the use of a hidden node, rather by another label from another task -- the \textsc{and} function, in this case. We highlight: there is no conditional label dependence at all between these two task labels wrt the true function, i.e., 
$P(y_\textsc{xor}|y_\textsc{and},\x) = P(y_\textsc{xor}|\x)$. 


Essentially, in view of neural networks: each label acts as an extra hidden node, and all $y_j\rightarrow y_k$ arrows represent additional trainable parameters, and each predictive model ($y_j$-node) is an activation function. The difference is that, rather than employing back propagation (which is not possible due to the model-agnostic assumption; the `activation functions' are classifiers though which we cannot back-propagate a gradient), we simply transfer predictive capacity from an existing trained function. Therefore, this is a kind of `deep' neural network (admittedly in this example, in the most minimalistic definition possible), without the deep learning. 

\subsubsection{Trading off variance for bias}
\label{sec:js}

Even several decades ago, the James-Stein estimator showed it was possible to benefit from modeling target variables together even if those variables are intrinsically independent from each other. This idea was taken up with regard to neural networks by \cite{ModularNeuralNetworks} but has only recently been revived in the analysis of multi-target learning by \cite{UnifiedView} (including multi-label and multi-task classification), mentioned specifically with regard to classifier chains by \cite{CCReview}, and to some extent in the deep-learning community \citep{copycats}, and also in the context of multi-task learning \citep{SteinMultiTask}. 

As \cite{UnifiedView} explain, from the machine learning viewpoint the issue is that of regularization. Modeling multiple problems together, instead of independently, means that parameters are shared, which in turn discourages overspecialization and overfitting. We develop this view further as follows. 

\begin{figure}[h!]
	\centering
	\includegraphics[width=0.35\textwidth]{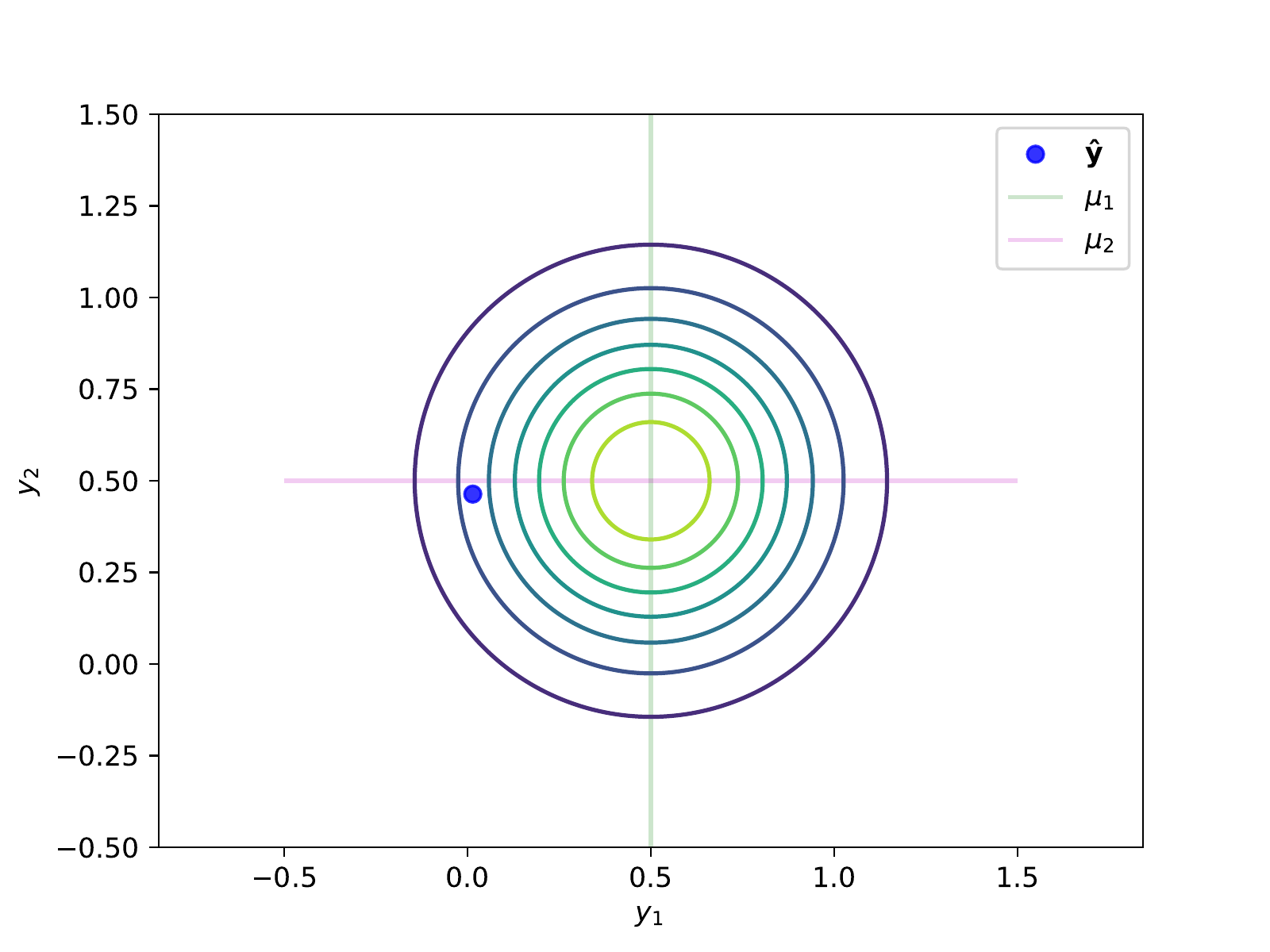}
	\caption{\label{fig:gausssian}The multi-label/multi-target problem illustrated for $m=2$ targets with a fictitious Gaussian to exemplify $P(\y|\x)$ of mean $\vmu = [\mu_1,\mu_2]$, and estimate thereof, $\ypred$. Note that $\vmu$ is not necessarily the true value of $\y$ (for this particular $\x$) due to irreducible error; $\y = \vmu + \error$. However, on average (in expectation) predicting $\vmu$ would give us the best results under a squared- or Hamming-loss metric. Since $\vmu$ is unknown, we build a model and come up with estimate $\ypred$; exemplified in the figure. 
	The shape of the Gaussian indicates no dependence among the two targets. Nevertheless, according to the James stein estimator it can be beneficial to model them together (at least, when $m>2$). 
	}
\end{figure}

 
Suppose that we want an estimate of $\vmu = \Exp[\Y|\x]$. This task is relevant in the multi-label regression context, where $\y \in \R^m$, because using it as a prediction would minimize loss metrics $\ell(\vmu,\y)$ based on squared-error, which are by far the most common in regression. And it is also relevant to multi-label classification; noting that $\vmu = \Exp[\Y|\x] = \sum_{\y} \y \cdot P(\y|\x)$, i.e., a hypothetical weighted vote over all combinations $\y \in \{0,1\}^m$; and predictions based on this vote, i.e., $\yp_j = \indicator{\mu_j \geq 0.5} \mid j=1,\ldots,m$ will indeed minimize Hamming loss. Fig.~\ref{fig:gausssian} illustrates. 

\begin{figure}[h!]
	\centering
	\subfigure[JS$_1$ properties: $\|\y\|^2$ vs $\lambda$ ($m=3,\sigma=1$)]{%
		\includegraphics[width=0.35\textwidth]{\basepath/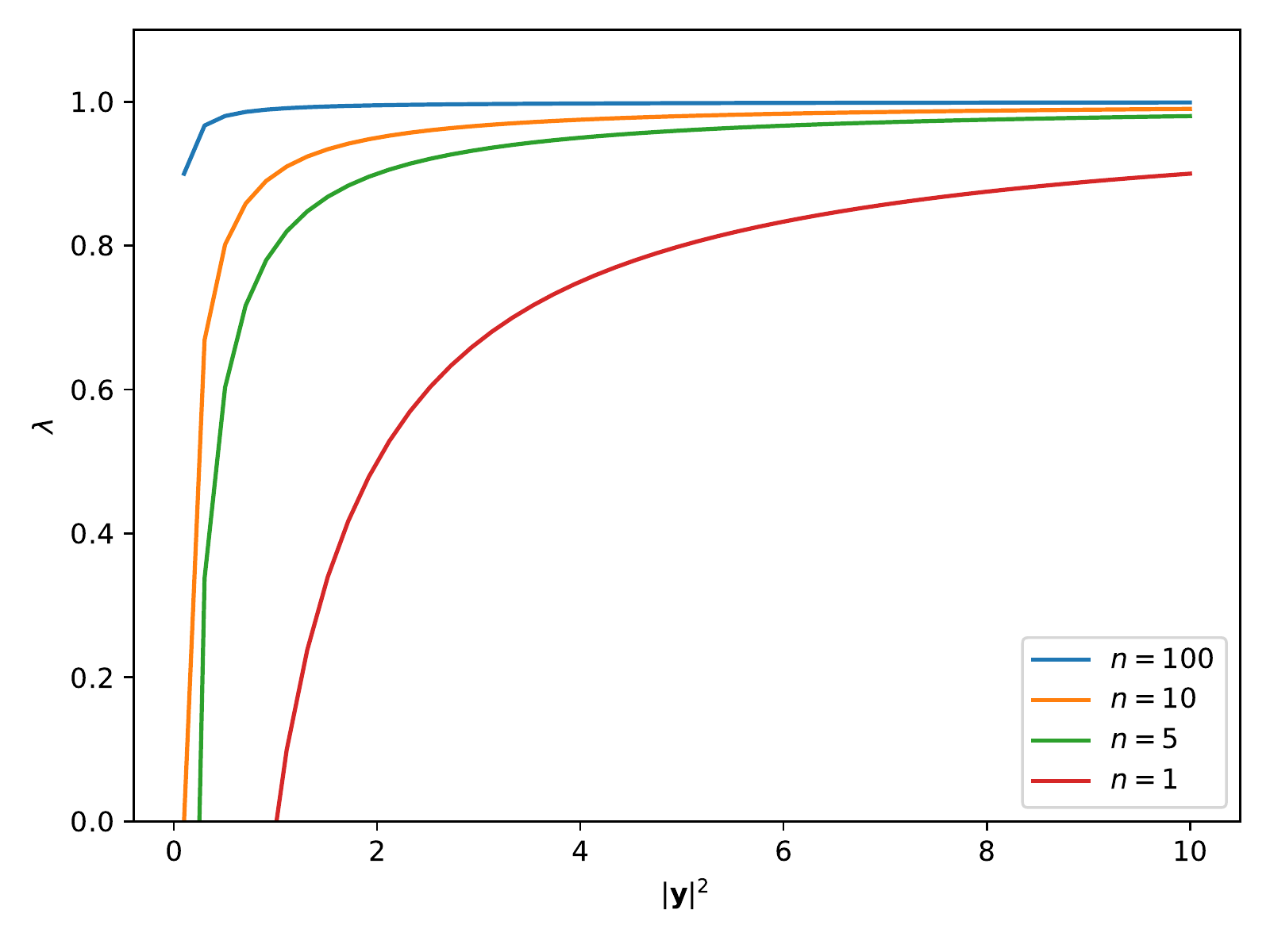}
		\label{fig:JS_a}
	}
	\subfigure[$\mathrm{JS}_1$ vs LS (wrt $n$; $m=5$); multi-target regression]{%
		\includegraphics[width=0.4\textwidth]{\basepath/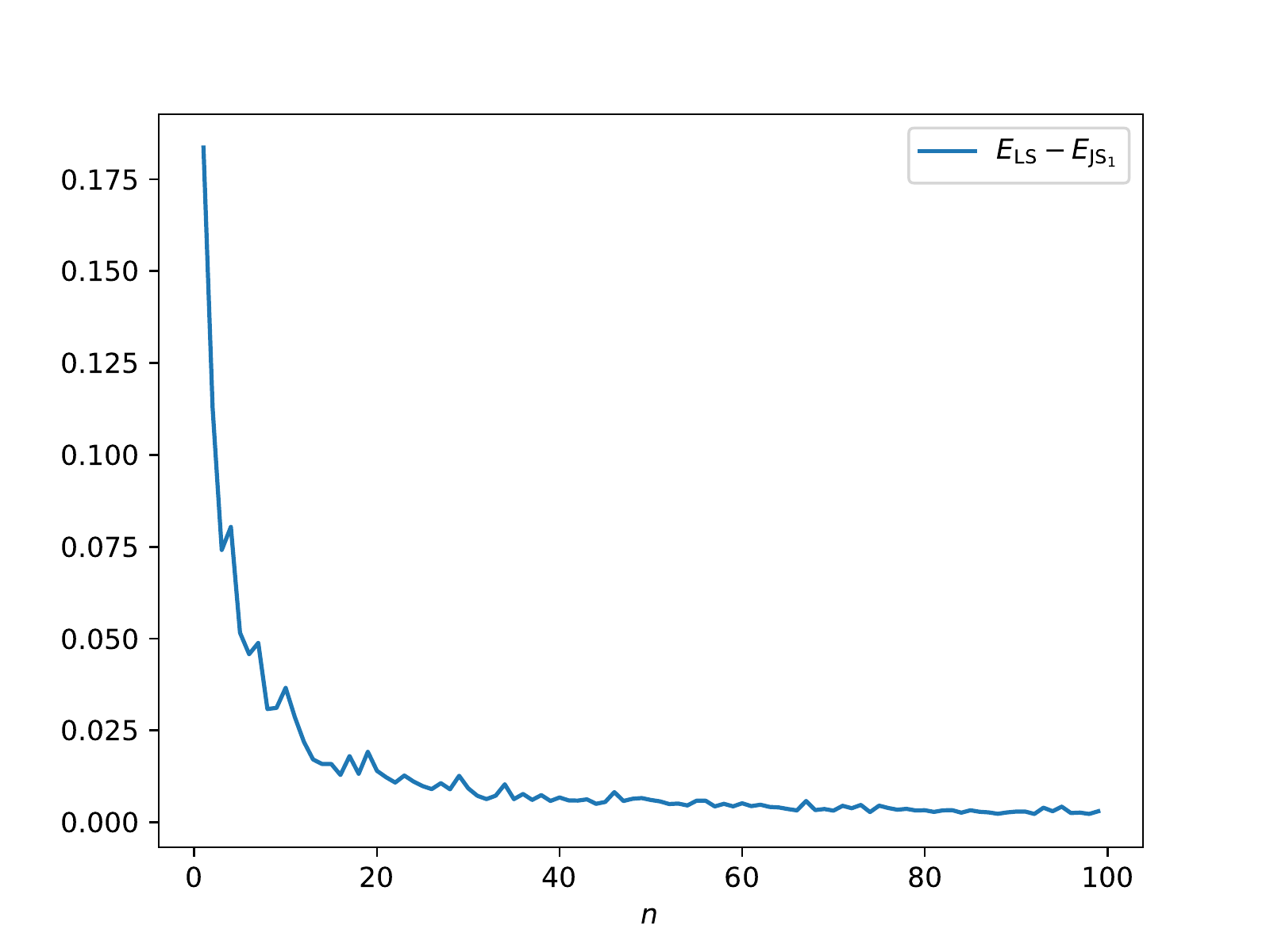}
		\label{fig:JS_b}
	}
	\subfigure[$\mathrm{JS}_1$ vs LS (wrt $m$; $n=30$); multi-label classification]{%
		\includegraphics[width=0.4\textwidth]{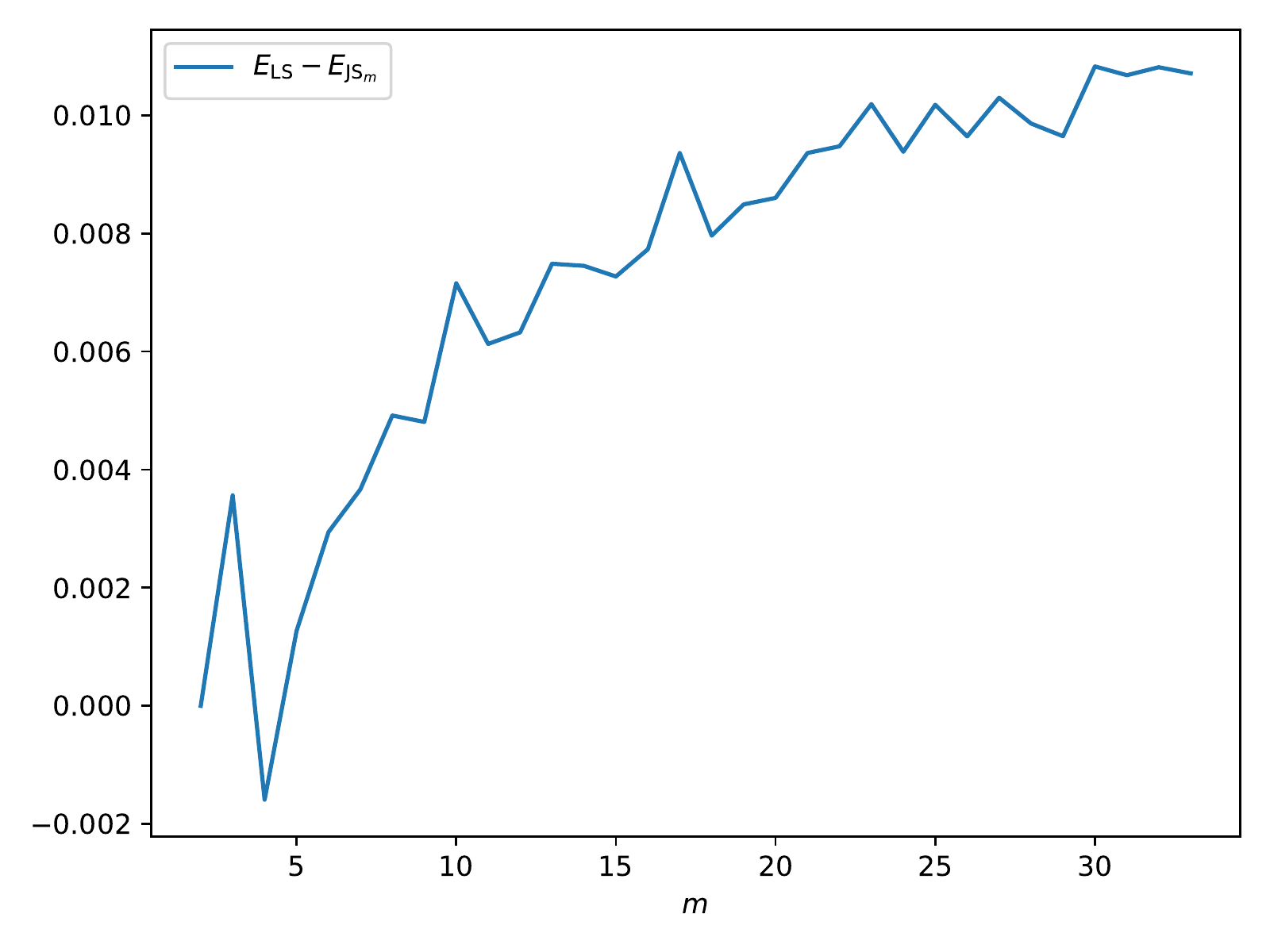}
		\label{fig:JS_c}
	}

	\caption{\label{fig:JS} Showing properties of \fref{fig:JS_a} regularization effect of James Stein (analytical), \fref{fig:JS_b} performance wrt number of training examples $n$ and \fref{fig:JS_c} wrt number of labels $m$ (empirical). Data is synthetic, with label concepts generated completely independently of each other. $E_A$ is error metric $E$ (mean squared error, in these figures) on algorithm $A$, where JS vs LS is James Stein vs Least Squares (maximum likelihood estimate). Note that higher values of $E_{LS} - E_{\mathsf{JS}_1}$ (improvement of JS over LS) is better. Each point has been generated from averaging over multiple runs. Stochastic gradient descent (SGD) is always the base learner (becoming logistic regression in \fref{fig:JS_c} for classification). 
	}
\end{figure}

Given model $h$ we can make multi-label decision $\ypred = h(\x)$, which we can think of as a sample from $P(\y|\x)$. We can also bootstrap from the dataset or, for example, randomize the structure of a joint-modeling method, thus obtaining multiple models from which we obtain  
multiple estimates/predictions $\{\ypred^{(t)}\}_{t=1}^n$\footnote{Note that samples $\y^{(t)} = [y_1,\ldots,y_m] \mid t=1,\ldots,n$ from a hypothetical posterior $\y^{(t)} \sim \Phat(\y|\x)$ should be distinguished from training samples $\y^{(i)} = [y_1,\ldots,y_m] \mid i = 1,\ldots,n$ from the data set, sourced originally as $\y^{(i)} \sim P(\y|\x)$, as denoted elsewhere}. We are essentially obtaining a number of samples \( \ypred^{(t)} \sim P(\y|\x) \mid t=1,\ldots,n \). A maximum likelihood estimate gives us unbiased estimate 
\begin{equation}
	\label{eq:mle}
	\ymean =  \frac{1}{n}  \sum_{t=1}^n \ypred^{(t)} 
\end{equation}
i.e., $\ymean \approx \vmu$. And (differently) the James-Stein estimator: 
\begin{align}
	\ypred_{\mathsf{JS}_n} &= \frac{1 - (m - 2) \frac{\sigest^2}{n}}{\|\ymean\|^2} \cdot \ymean \label{eq:1} \\
				&= \lambda(\{\y^{(t)}\}) \cdot \ymean \label{eq:2}
\end{align}
where we have explicitly denoted $\lambda(\cdot)$ as a mechanism to regularize $\ymean$, pulling it towards $0$ whenever $1-(m-2)\frac{\sigest^2}{n} < \|\ymean\|$; shown graphically in \Fig{fig:JS_a}. Note that from $\{\y^{(t)}\}_{t=1}^n$ we have $n,\sigest,m,\ymean$ (enough to fully express \Eq{eq:1}). 

Pulling estimates towards $0$ makes particular sense in the multi-label classification case 
because label vectors are relatively sparse. In other words, for any given label $j$, the expected label $\Exp[Y_j]$ is indeed more likely to be closer to $0$ than to $1$. 
When modeling independent labels together, we impute this information in the form of a biased estimator.



\Fig{fig:JS_b} and Fig.~\ref{fig:JS_c} confirm the potential for empirical gains by modeling unrelated labels together (via the James-Stein estimator) even though they are generated independently from each other. However, we also observe that these gains are very slight, except for large $m$ and, more particularly, very small $n$. 
Which confirms earlier speculation (e.g., by \cite{UnifiedView}). 


\subsubsection{The multi-label ensemble effect}

Many multi-label methods are developed and presented in the context of an ensemble, then compared successfully to independent models. Success, measured as relative accuracy gain, is usually attributed to the design of the novel method component presented, and overall the proposed scheme of modeling label dependence. However, the effect of using an ensemble (vs single) model could, in many cases, be a significant component of such an accuracy; i.e., some gains are obtained regardless of the existence (or not) of dependence between the task labels. Essentially: if multi-label method X is presented in terms of `ensembles of X' (a common phrasing in the literature), we cannot easily distinguish how effective X is, unless presented alongside ensembles of independent models; illustrated in Fig.~\ref{fig:stacking}. Indeed, ensembles of independent models have been recognized as competitive \citep{ExtMLEnsemble}. 

It is true that adding more models (to an ensemble) also provides the opportunity to include more predictive power (discussed in \Sec{sec:non_lin}), as well as take into account more label dependence (discussed in Section~\ref{sec:label_dependence}), when making predictions. To unravel this overlap we carry out a study to specifically isolate the ensemble effect: first, we generate totally-independent regression tasks $\dX_j \rightarrow \dY_j$ ($\forall j=1,\ldots,m$)  then append them together as if a single multi-output problem; and then we experiment with independent linear regression models vs a chain models of linear regression models vs ensembles of each of these. It is already established that extra capacity cannot be provided by the chains/joint modeling \citep{HanenMORSurvey} (because without non-linearities, the structure collapses); 
and similarly, we can easily see this is also the case in the multi-output ensemble: 
in the ensemble case, with regression weights $\w$ and ensemble averaging process (across $n$ models, as above in Section~\ref{sec:js}): 
\begin{align*}
	\ypred &= \frac{1}{n} \left\{ \x\w^{(1)} + \cdots + \x\w^{(t)} + \cdots+ \x\w^{(n)} \right\} \\
			&= \frac{1}{n} \x ( \w^{(1)} + \cdots + \w^{(t)} + \cdots + \w^{(n)} )  \\
			&= \w \x 
\end{align*}
where, let $\w/n = \w^{(1)} + \cdots + \w^{(n)}$. 

\begin{figure}[h!]
	\centering
	\includegraphics[width=0.4\textwidth]{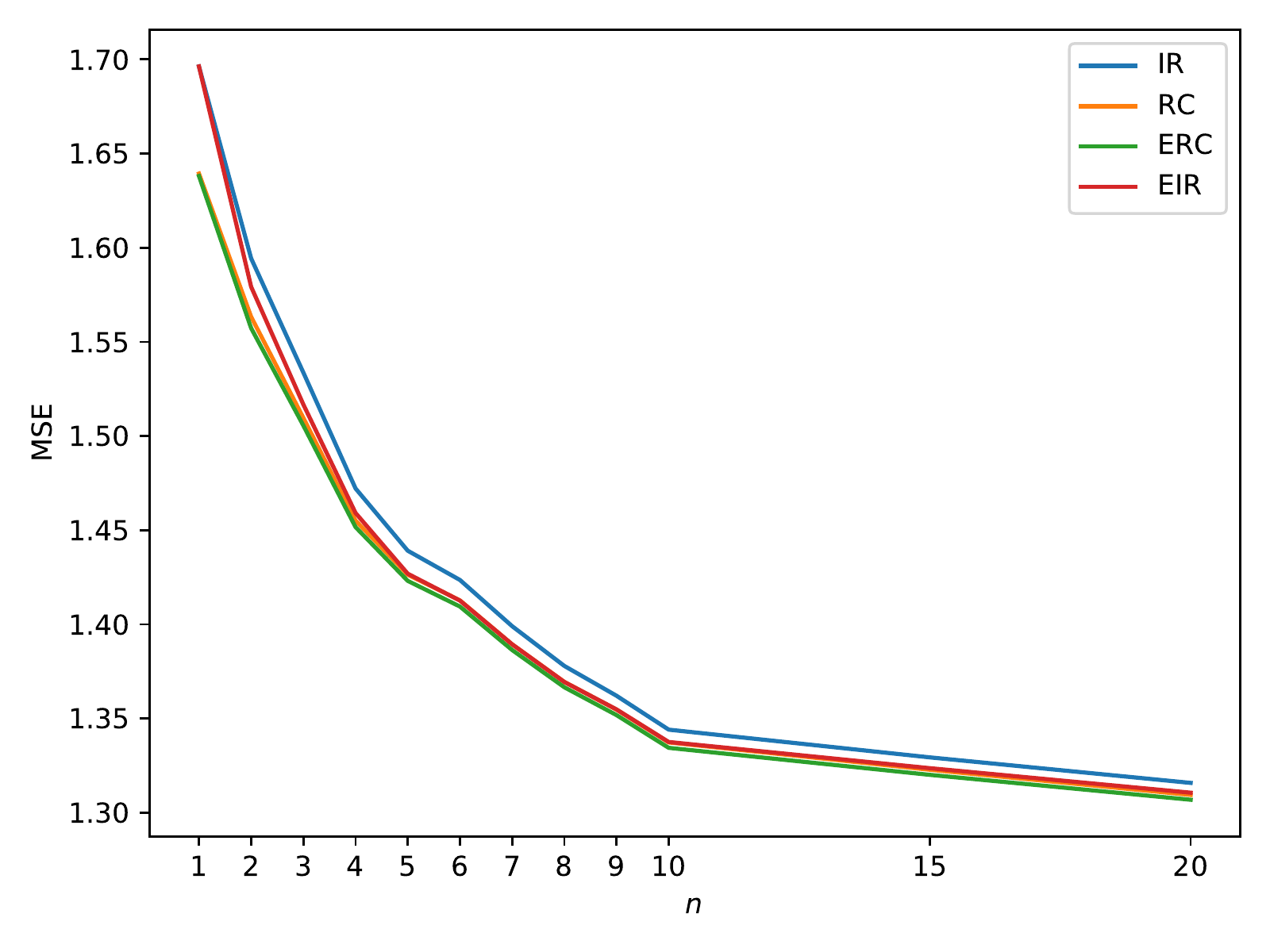}
	\caption{\label{fig:JS_d}
		The mean squared error (MSE) of independent regression models (\textsf{IR}) vs regressor chains (\textsf{RC}) vs ensembles of regressor chains (\textsf{ERC}; each chain a random order) vs ensembles of independent regression models (\textsf{EIR}; standard bagging ensemble) wrt number of  $n$, where $m=10$. Hence, demonstrating the `ensemble effect', which is slight, but visible. Data generation and general setup is equivalent to \Fig{fig:JS}; each point averaged over 100 simulations. 
	}
\end{figure}

Therefore, the remaining improvement in error reduction should indicate regularization in the context of variance reduction, i.e., the `ensemble effect'. Results are given in Fig.~\ref{fig:JS_d}. Indeed, despite that none of the four methods have any theoretical advantage among them in terms of model capacity or the ability to model dependence (which is non existent in this case), there is a non-negligible difference in accuracy. We can also discard bias-reduction such as implied by, e.g., the James-stein estimator, since we are considering only \Eq{eq:mle}. 

Vis-a-vis the James Stein estimator, this `ensemble effect' concerns only the non-bias estimate in Eq.~\eqref{eq:mle} (which appears in \Eq{eq:2}). It is known to be the reduction of the variance component of the error, achieved by averaging votes of different estimators of the same target variable. For example, the well-established technique of bootstrap aggregating (bagging) \cite{Bagging}. This effect is not specific to the context of a multi-output problem, unlike the James-Stein estimator, which will add bias to an estimate via $\lambda(\cdot)$ by considering the existence of other labels. 
Improvement can only be achieved on instability in the instantiation of base learner. Chained models produce this instability simply with random chain orders for each base learner which is known to produce ample diversity \citep{CCReview}, and thus the actual bagging procedure (of sampling the training data) is not needed. This is apparently (according to the results in Fig.~\ref{fig:JS_d} and the discussion herein) even the case when there is no dependence for such a chain to model; features in that case provide noise, which provide necessary diversity for ensemble voting to work. 

\subsubsection{Stacking, error propagation and correction} 

\begin{figure}[h!]
	\centering
	\subfigure[Multi-label Ensemble]{%
		\label{fig:bagging}
	\includegraphics[scale=1]{\basepath/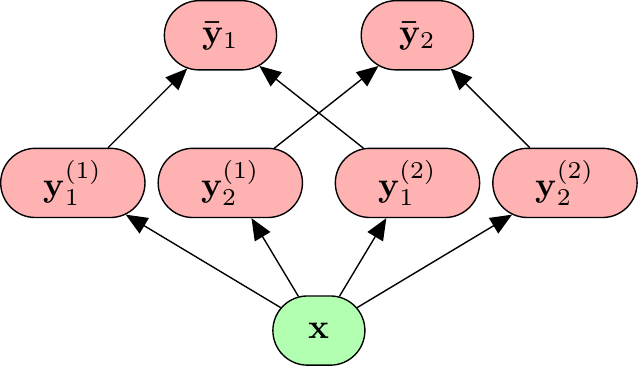}
	}
	\quad
	\subfigure[Multi-label Stacking]{%
		\label{fig:stacking}
	\includegraphics[scale=1]{\basepath/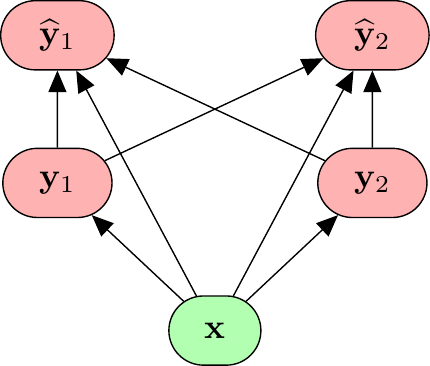}
	}
	\caption{\label{fig:stacking_vs_bagging}A \fref{fig:bagging} bagging ensemble (of $n=2$ \emph{independent} models/label concepts; not affecting each other's predictions) and \fref{fig:stacking} stacked prediction (also $n=2$). We have kept the multi-dimensional label generalization in line with, e.g., Fig.~\ref{fig:the_two}; i.e., two multi-dimensional label concepts ($m=1$); with an additional glyph over $\y$s of the second label layer -- to distinguish from the first.}
\end{figure}

There is a second mechanism at play in Fig.~\ref{fig:JS_d}.	Chained models (\textsf{RC}) also provide improvement wrt independent regressors (\textsf{RC}) -- despite not providing any advantage in either dependency-modeling or predictive capacity. This is a different mechanism than vanilla ensembles. Whereas bagging expands horizontally across $n$ models and then averages their results per label; in chained models, predictions are `adjusted' wrt one another, in a vertical fashion of two ore more levels. This is similar to the general concept of stacking \citep{EOSL}; where one prediction can affect another. Both approaches are contrasted in Fig.~\ref{fig:stacking_vs_bagging} in the multi-target context. Note that whereas in the vanilla ensemble averaging, predictions for a label/task $j$ are averaged together (i.e., all paths leading to $\ybar_j$ pass only through $\{y_j^{(t)}\}_{t=1}^n$), and this is not the case in stacked models. 

This mechanism of layering predictions (such as that of Fig.~\ref{fig:stacking}, but there are many similar varieties) has been credited for improvement over independent models several times in the multi-label literature, e.g., employed in \citep{IBLR,MLCStacking2}, recently discussed by \cite{UnifiedView} in a general multi-output context, and specifically in the context of chaining by \cite{NestedStacking}. However, although stacking is known generally in machine learning, the presence of multiple labels/tasks creates a different dynamic where label predictions are corrected based on earlier predictions from \emph{different} labels. In some contexts it is called error correction (or, equivalently claimed to reduce error propagation) but the mechanism being referred to is the same. 

In reflection of the earlier development of the James-Stein estimator in Eq.~\eqref{eq:2}, whereas JS biases/pulls an otherwise non-biased estimate $\ymean$ according to term $\lambda(\cdot)$, bagging focusses only on the $\ymean$ term, and stacking focusses only only $\lambda(\cdot)$. 
The relationship between the different methodologies can be expressed in the following among multi-label models $h^{(t)} : \dX \rightarrow \R^m \mid t =  1 ,\ldots,n$: 
\begin{align*}
	\ypred^{(t)}  &= h^{(t)}(\x) = [y_1^{(t)}, \ldots, y_m^{(t)}] \\
	\ymean &=  \frac{1}{n} \sum_{t=1}^n \ypred^{(t)} = \left[ \sum_{t=1}^n y^{(t)}_1, \ldots, \sum_{t=1}^n y^{(t)}_m \right] \\
	\ypred_\mathsf{JS_n} &= \lambda(\{\ypred^{(t)}\}) \cdot \ymean \\
	\ypred_\mathsf{JS_1} &= \lambda(\{\ypred^{(1)}\}) \cdot \ypred^{(1)} \\
						 &= \lambda(\{h^{(1)}(\x)\}) \cdot h^{(1)}(\x) \\
	\ypred_\mathsf{Stacking} &= \lambda(\x, h^{(1)}(\x)) 
\end{align*}
where $\lambda(\cdot)$ is, e.g., a multi-label meta model, such as the ones mapping $\lambda : \y_j \times \x \mapsto \ypred_j$ in Fig.~\ref{fig:stacking}.

How does such a mechanism (stacking) work if there is no label dependence? Although small, it is worth pointing out that the gain of using a regressor chain (no ensemble) offers already most of the gain in Fig.~\ref{fig:JS_d}.

A single chain model can also be viewed as a particular variety of stacking. However, it is worth mentioning (as also discussed by \cite{NestedStacking,UnifiedView,CCReview}) here that stacking is not necessarily equivalent to chaining due to a potentially different training procedure concerning the source of the training label vectors $\y$s. In stacking, second-level models $\lambda$ draw training data from the first layer models thus $\y \sim \Phat(\y|\x)$ where $\Phat$ characterises the underlying probabilistic interpretation of the model $h$. In training chains, labels are lifted directly from the training data, i.e., sourced from the distribution $\y \sim P(\y|\x)$. 
So, in vanilla stacking, the second layer $\lambda$ reduces bias of the \emph{individual} models $h_j$, but at a cost of eliminating bias in the form of label combinations (i.e., at the cost of not modelling label dependence), and thus again the effect is mostly regularization via bias reduction, rather than benefiting explicitly from label dependence. This is confirmed by \cite{DCC2,NestedStacking} and others who empirically find that gains vs independent models were indeed under Hamming loss (which does not explicitly benefit from any model of existing label dependence) rather than $0/1$-loss (which does). 




Specific to model chains, the concept of \emph{error propagation} has been evoked numerous times, e.g., \citep{CCPropagation} (and references therein). This refers to the idea that a poor prediction will cascade along the chain and negatively influence later predictions (i.e., more poor predictions for different labels). Interestingly, this is exactly the opposite affect that stacking is supposed to achieve (of \emph{correcting} predictions from one label to another) and thus highlights the complexities induced by the multi-output chaining model. Obviously, one seeks to \emph{minimize} the propagation of error along a chain of predictions. 

Error propagation certainly exists under a probabilistic interpretation of chaining when modeling label dependence for the minimization of non-decomposable loss metrics such as $0/1$ loss (indeed, as shown by \cite{CCPropagation} under the term of `label noise', caused by training labels coming from a different distribution to the true labels). 
However, in the feed-forward context, we argue that the idea of minimizing error propagation contradicts with the idea of providing predictive capacity to models further down the chain (discussed above in \Sec{sec:non_lin}). One does not study whether an activation function of a neural network provides a `correct prediction' in order to be a useful feature to the next neuron. Empirical investigations also offer little to no support for this effect of error-propagation reduction: putting easier/reliable labels first does not appear to increase chain performance \citep{CCReview} (and references therein). 
And most importantly, we note that the probabilistic minimization is not relevant to transfer learning; where we are only interested and in control of the performance of the second (target) task. 
 


\subsection{A study on effect interaction}
\label{sec:interaction}

Having so far discussed the hypothetical benefits of modeling labels (i.e., tasks) together even when they bear no measurable statistical dependence among each other, the question we ask now is -- to what extent these effects can be translated into gains on real-world and benchmark multi-label datasets? (in view of translating those effects into cross-domain transfer learning).  

\begin{figure}[h!]
	\centering
	\subfigure[ ]{%
		 \includegraphics[width=0.4\textwidth]{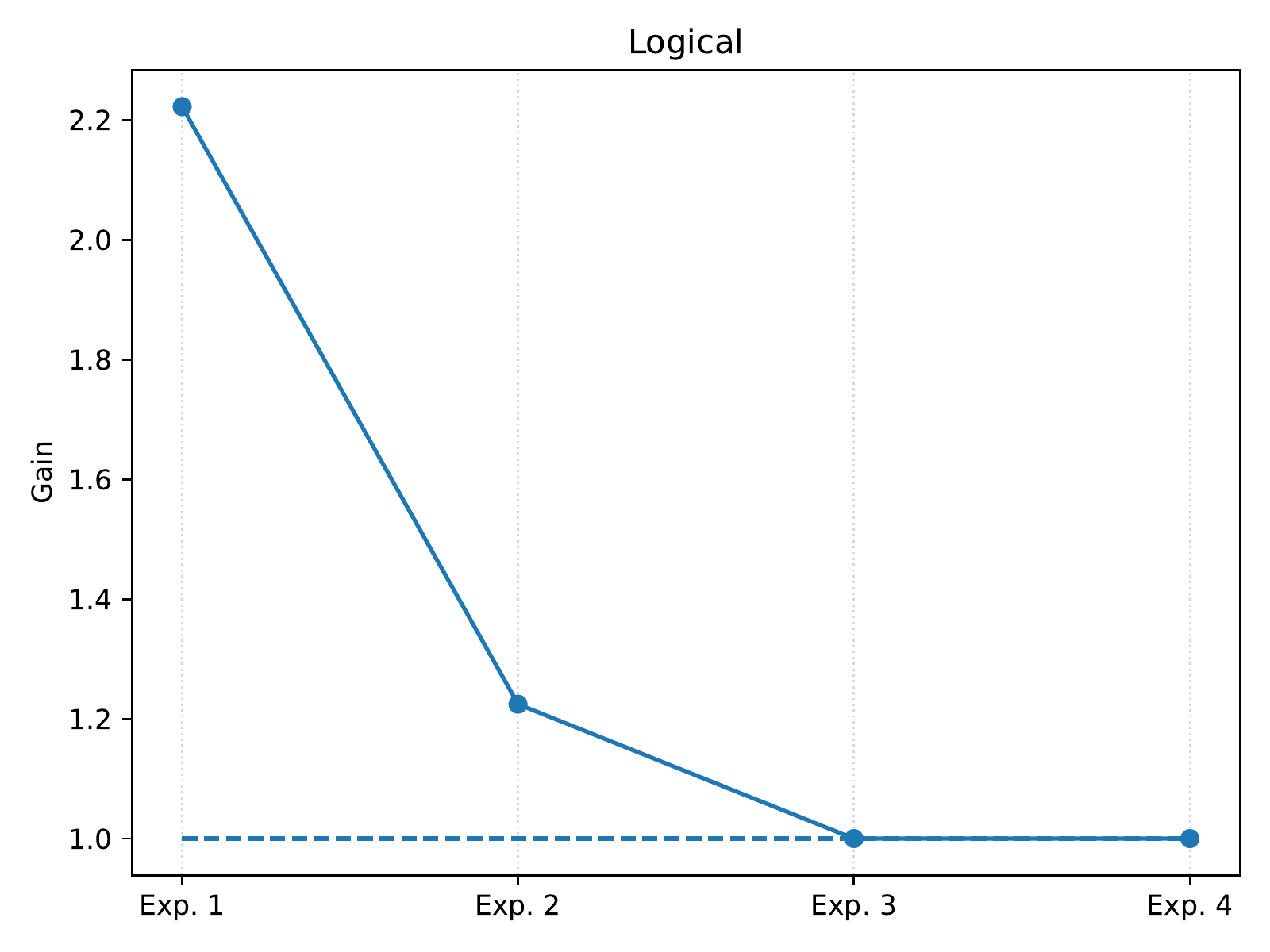}  
	 }
	\subfigure[ ]{%
		\includegraphics[width=0.4\textwidth]{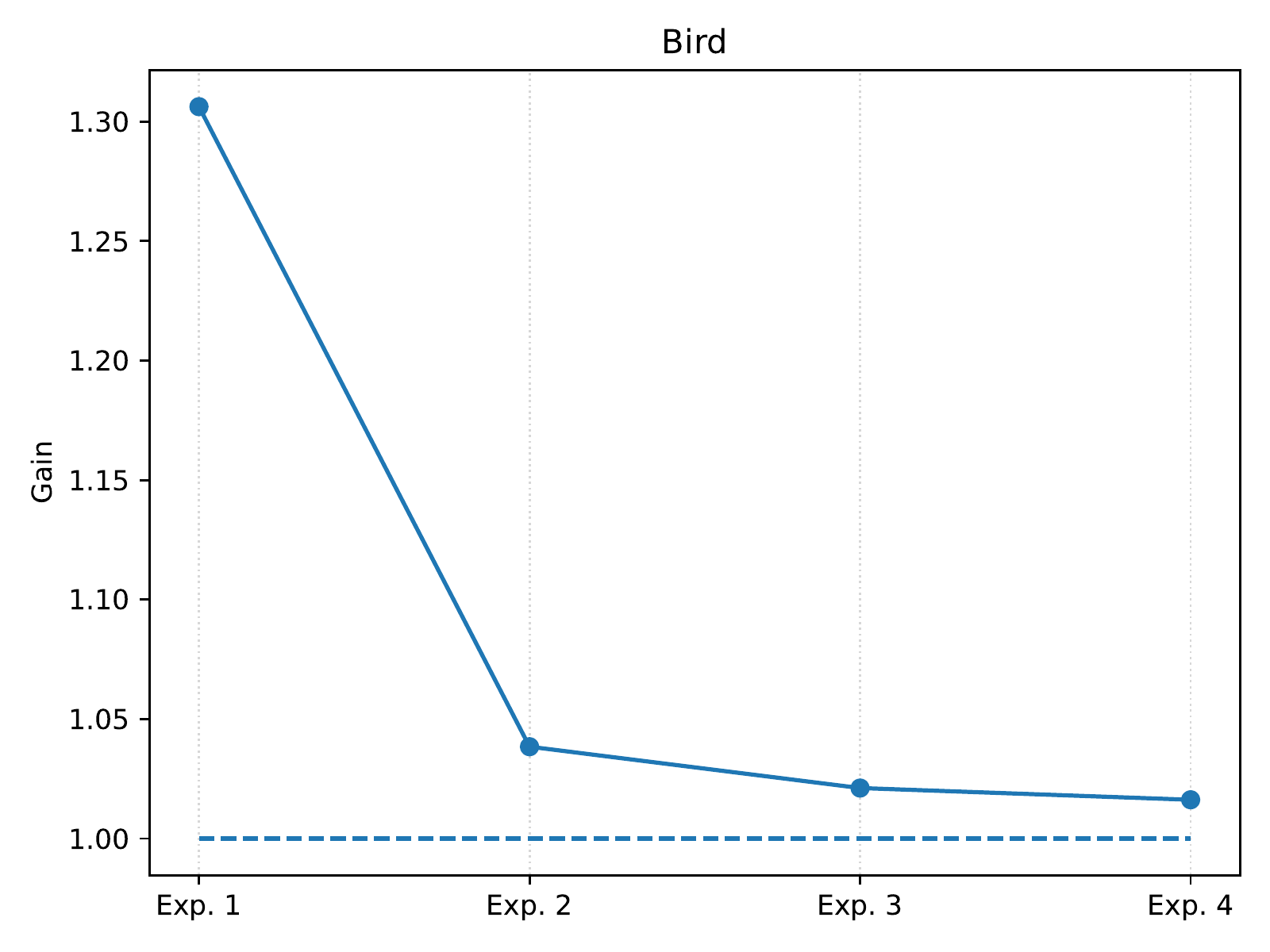}  
	}\\
	\subfigure[ ]{%
		\includegraphics[width=0.4\textwidth]{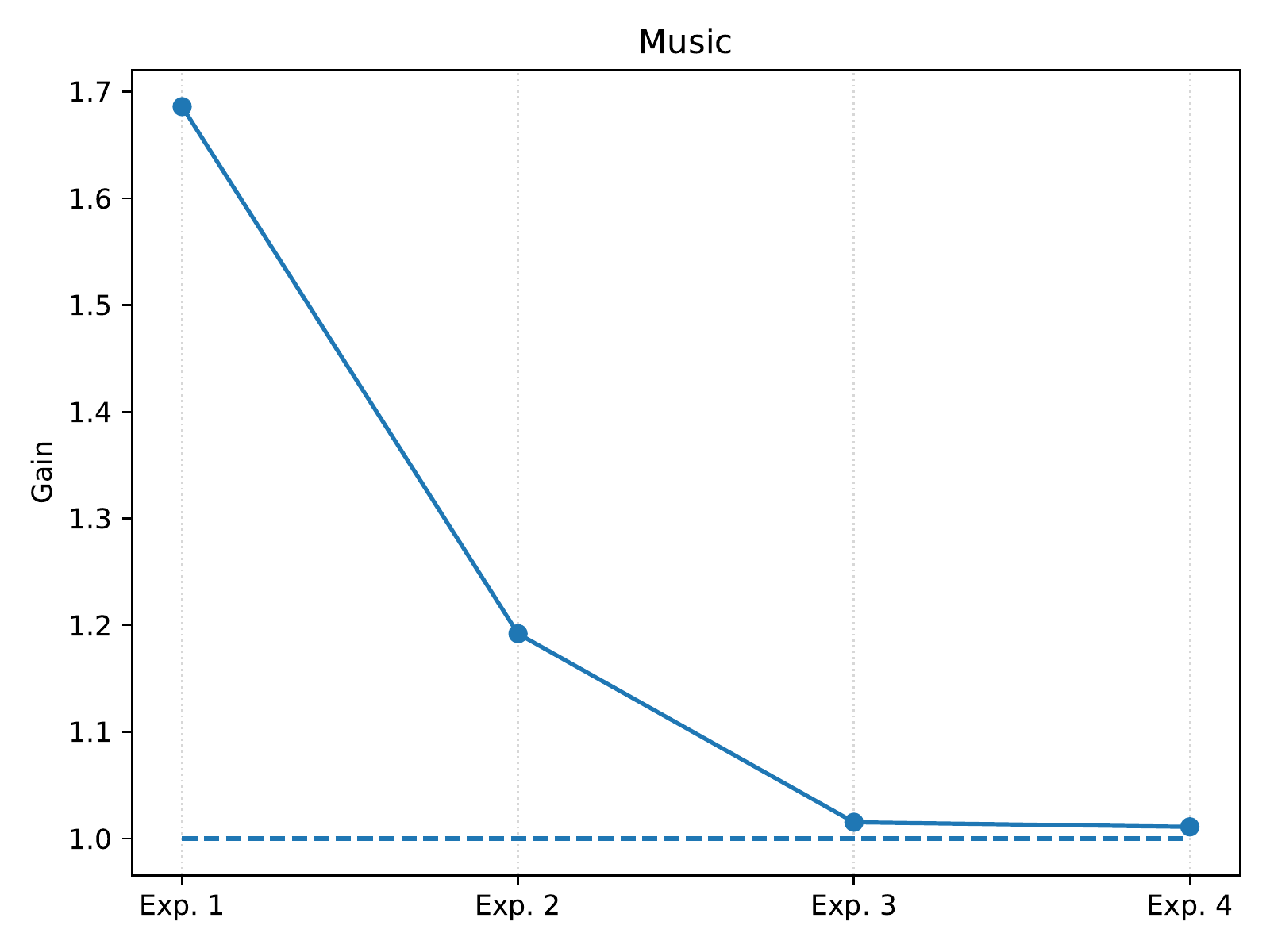}  
	}
	\subfigure[ ]{%
		\includegraphics[width=0.4\textwidth]{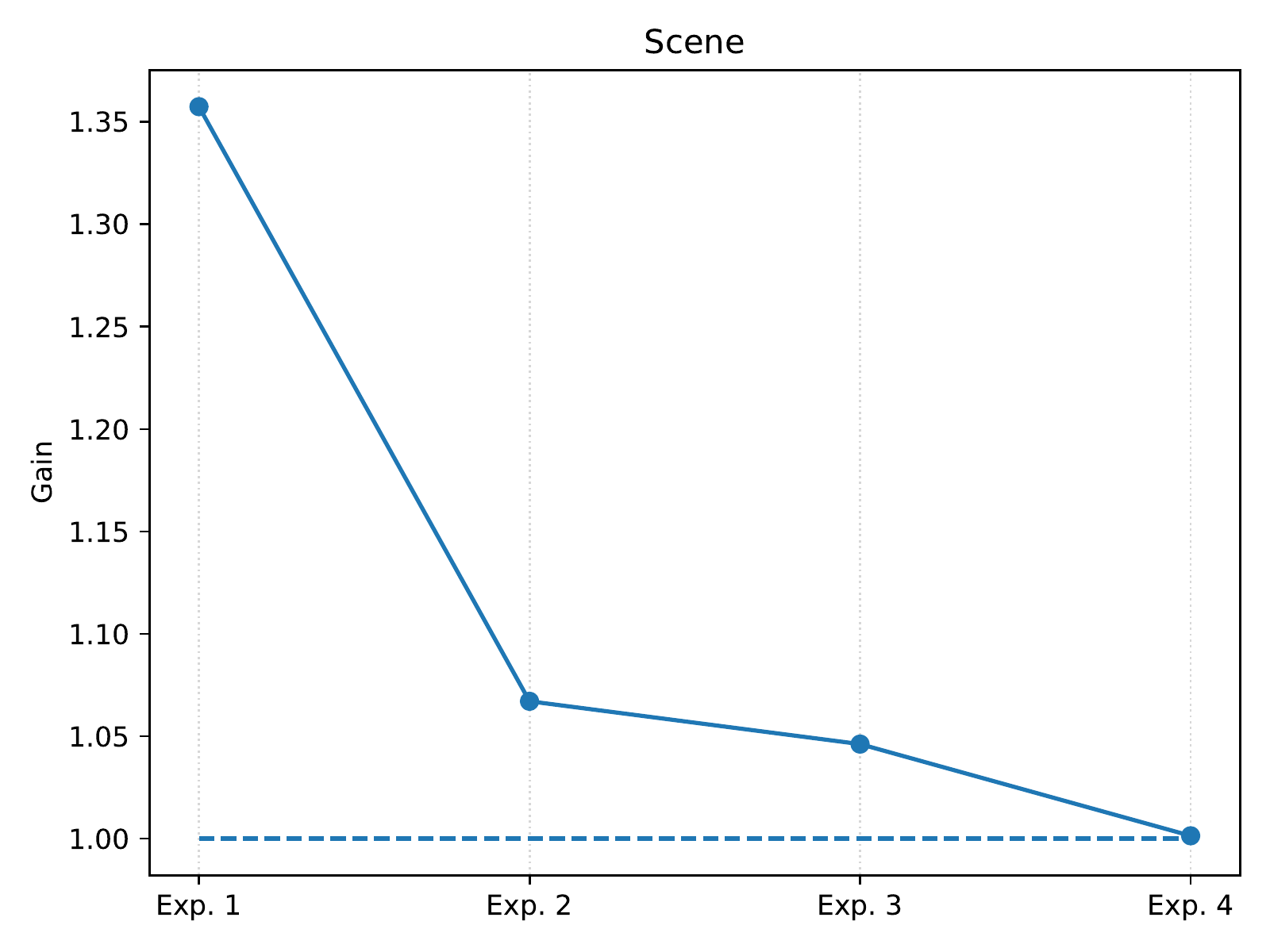} 
		\label{fig:scen}
	} \\   
	\subfigure[ ]{%
		\includegraphics[width=0.4\textwidth]{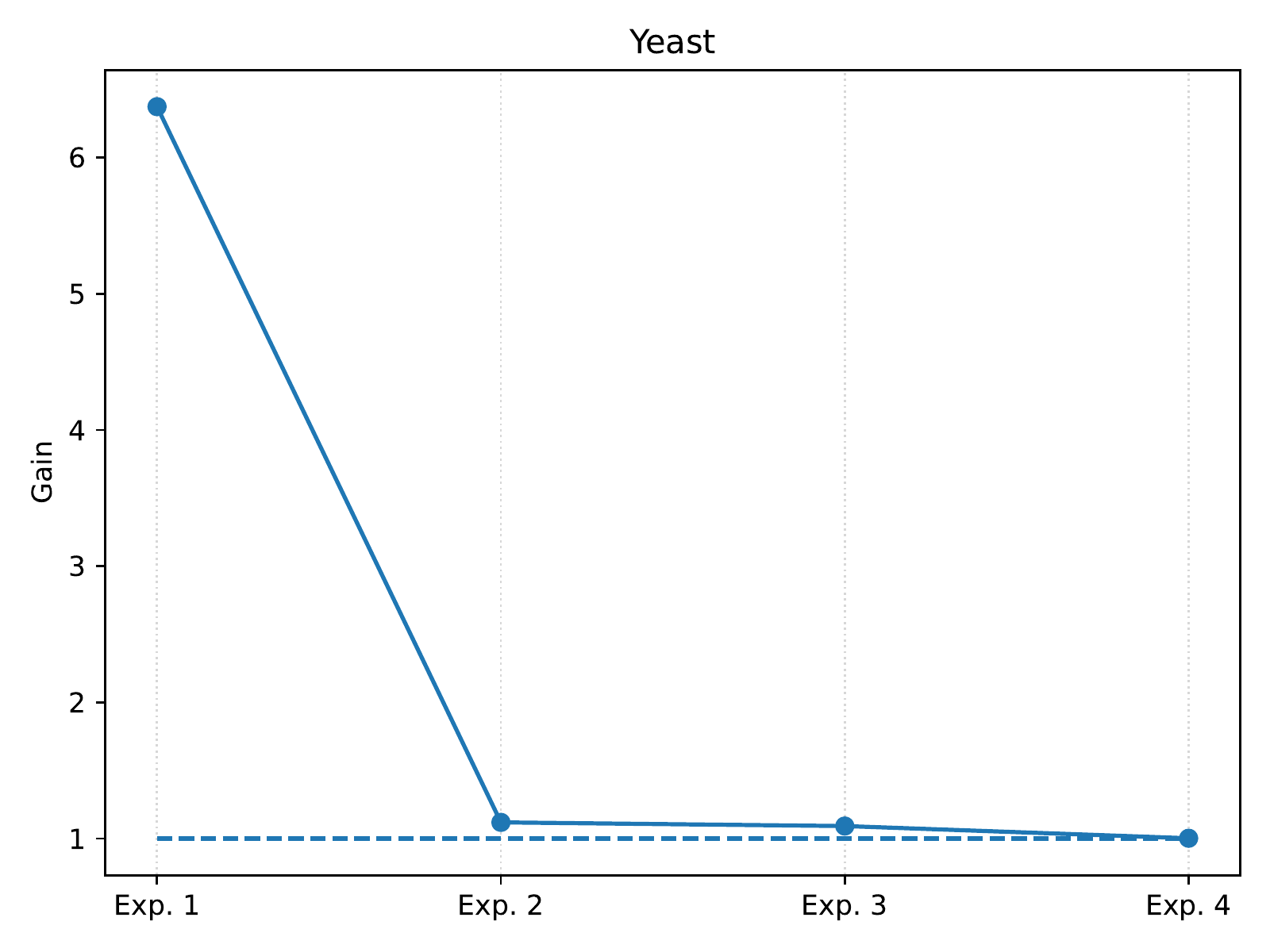} 
	}
	\caption{\label{fig:chain_gain} The gain of modeling labels together (across a cascade/chain) over independent models on well-known benchmark multi-label data sets (listed in Table~\ref{tab:datasets}); indicating how many times better the predictive performance is:  
	\(
		\textsf{Gain} = \frac{1 - \ell(\text{chains})}{1 - \ell(\text{indep.})}  
	\)
	where $\ell(h) \in [0,1]$ is the loss obtained by employing model $h$. Therefore a gain of 1.2 indicates that a chain-based model obtains 20\% better predictive performance that independent models applied on the same problem.  
	Each point is obtained as an average over 5-fold cross validation, on experiments 1.--4.\ as listed above. }
\end{figure}

In order to investigate, we repeat the following sequence of experiments on several standard multi-label datasets; results displayed in \Fig{fig:chain_gain} (experimental details such as framework and the datasets are described in Section~\ref{sec:experiments}). The effects that we expect to capture are denoted in bold.  

\begin{enumerate}
	\item [Exp 1.] \textbf{Models of label dependence, non-linearity, regularization}: We compare independent models of vanilla logistic regression vs an ensemble of classifier chains with logistic regression as base model, evaluated under the $0/1$ loss metric (\Eq{eq:01}). 
	\item[Exp 2.] \textbf{Non-linearity, regularization}: As above, except we change the loss metric to a decomposable one: Hamming loss (\Eq{eq:HL}); for which no model of label dependence is required for optimum performance. Therefore any difference in performance must be explained by other mechanisms, namely extra capacity/non-linearity (given in the form of the classifier chain\footnote{Logistic regression provides a linear decision boundary, a non-linearity is produced by converting the regression score $\in [0,1]$ into  label $\in \{0,1\}$}) and regularization thereof; with variance reduction possible via both ensembles, and bias-reduction possibly via the chain structure among models.  
		In other words: models of label-dependence should play no role here. 
	\item [Exp 3.] \textbf{Regularization:} As above, except now replacing logistic regression as a base classifier with multi-layer perceptrons (MLP) in both cases. MLPs provide sufficient architecture for non-linearities, so any additional capacity provided by the classifier chains should be superfluous. At this point the only gain of the ensemble of classifier chains (vs independent models) must be regularization. 
	\item [Exp 4.] \textbf{Spurious effects only:} As above but we now add regularization (in the form of a typical L2 weight penalty) to all MLP instantiations. 
		According to our hypotheses, any gain of chaining found here should be treated as spurious; not likely to be replicated in general. 
\end{enumerate}


Results in \Fig{fig:chain_gain} confirm the widely-held belief in the literature that modeling label dependence is a core issue in multi-label learning (depending on which loss metric is under consideration), and support the idea that studies and models of such dependence can lead to strong gains over a benchmark method that treats each label independently. However, in addition, as we hypothesized, other techniques can have a non-negligible influence. Consider the \textsf{Music} dataset (\Fig{fig:scen}), one of the most widely-used benchmarks in hundreds of multi-label papers: effects \emph{other} than label-dependence (e.g., additional architecture, and regularization) have produced on average a model around 1.2 times more performant in terms of accuracy than a typical baseline of independent models. This is highly significant.  
Even gains of 1.05 times the benchmark are very interesting results when there is no reason otherwise to expect any benefit at all from label dependence.  The `spurious effects' of Experiment 4 under \textsf{Bird} should not be regarded necessarily as suspicious. The classifier chain still represents a different model, and although theoretically the architecture of an MLP should be powerfully non-linear, for example, the additional neurons of prior nodes are small in number but may be qualitatively useful as opposed to the generic and default activation functions. 

In the remainder of this paper, we look to translate these results to the domain of transfer learning. 


\section{Transfer Learning without Task Similarity via Transfer Chains}
\label{sec:development}

We infer from the results of the previous section that task similarity is only one of several aspects to consider when modeling multiple tasks together. In this section we develop the setting of model-driven model-agnostic (black-box) cross-domain transfer by removing precisely the assumption on task similarity, and nevertheless specifically leveraging the chaining mechanism. In this section we outline, justify, and explain our approach of `transfer chains', with some empirical experiments to examine its performance. 


\subsection{An outline of Transfer Chains} 
\label{sec:our_approach}

\begin{figure}[h!]
	\centering
	\subfigure[Transfer Chains]{%
		\includegraphics[scale=1]{\basepath/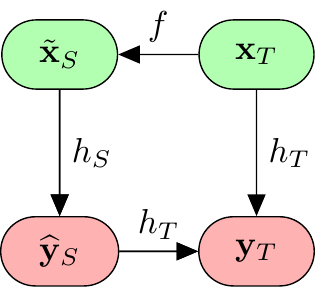}  
		\label{fig:tl_novel_0}
	}\quad\quad\quad
	\subfigure[Non-Transfer]{%
		\includegraphics[scale=1]{\basepath/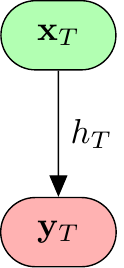}  
		\label{fig:tl_novel_non}
	}
	\caption{\label{fig:tl_novel_vs_not}A simplified schema of  \fref{fig:tl_novel_0} our proposed approach vs \fref{fig:tl_novel_non} traditional data-driven approach that does not make use of external models. Edge labels are explained via Eqs.~\eqref{eq:e1}--\eqref{eq:e3}. In particular, note that the connection from $\x_T \rightarrow \xtest_S$ implies an imputation/mapping step of $\x_T$ into $\dX_S$-space; and does not involve source-task data at any point. The main difference from \Fig{fig:intro} is that this is the model-agnostic approach where we cannot observe hidden layers.}
\end{figure}

\Fig{fig:tl_novel_0} depicts a simplified representation of our proposal. 
Our goal is to produce a target prediction $\ypred_T$ for given target test instance $\x_T$ (we look specifically at the classification case). We aim to make use of a source model $h_S$, but we have no access to any of its source training data pairs $(\x_S,\y_S)$; and we specifically denote $\xtest_S$ and $\ypred_S$ (with tilde, and hat, respectively for the input and output) to remind of this. 
The pipeline as follows (for notational simplicity, but without loss of generality, we depict scalar output variables rather than vectors for the remainder of this section): 
\begin{align}
	\xtest_S &= f(\x_T) \label{eq:e1} \\
	\yp_S &= h_S(\xtest_S) = \argmax_{y \in \dY_S} \Phat(y | \xtest_S) \label{eq:e2} \\
	\yp_T &= h_T(\yp_S,\x_T) = \argmax_{y \in \dY_T} \Phat(y | \x_T, \yp_S) \label{eq:e3} 
\end{align}
where $h_S$ and $h_T$ are the source and target task models, respectively; and $f$ is a function mapping from the source input to target input space:
\[
	f : \dX_T \rightarrow \dX_S
\]

The probabilistic interpretations $\Phat$ implied by models $h$ are not strictly part of the framework, but will be used to discuss insights from a statistical perspective in the following. They could be a parametric distribution such as in logistic regression or a distribution of votes as in, for example, a random forest. 

%



\subsection{Theoretical insights: manufactured dependence}

\begin{figure}[h!]
	\centering
	\subfigure[$Y_T \sim P(\cdot | \x_T)$]{%
		\includegraphics[scale=1]{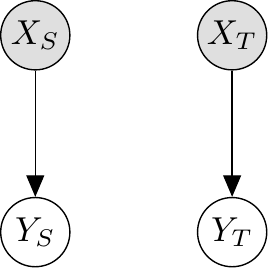} 
		\label{fig:t0}
	}\quad\quad\quad
	\subfigure[$\Yp_T \sim \Phat(\cdot | \x_T,\yp_S)$]{%
		\includegraphics[scale=1]{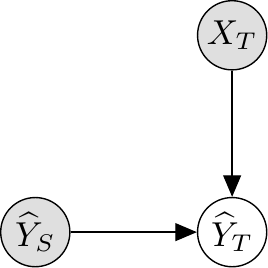}
		\label{fig:t1bis}
	}\quad\quad\quad
	\subfigure[]
	{%
 		\begin{minipage}{3cm}
 	  \begin{tabular}{llll}
 		  \hline
 			  $X_1$ & $X_2$ & $Y_{\textsc{and}}$ & $Y_{\textsc{xor}}$      \\
 		  \hline                                 
 			  0     & 0     & 0                  & 0                     \\
 			  0     & 1     & 1                  & 1                     \\
 			  1     & 0     & 1                  & 1                     \\
			  1     & 1     & 1                  & {\color{red}1} \\
 		  \hline
 	  \end{tabular}
 		\end{minipage}
 		\label{fig:modelB}
		}
	\caption{\label{fig:pgms}Probabilistic graphical models illustrating \fref{fig:t0} the generating distribution of $Y_T$, and \fref{fig:t1bis} the approximated predictive distribution of $\Yp_T$. An example \fref{fig:modelB} illustrates where this conditional dependence arises due to insufficient capacity of the base model; extended from Fig.~\ref{tab:xor_1}: supposing a linear base model class $h_T$, conditional dependence (and thus, a motivation to model these tasks together) has been created (the affected bit is highlighted in {\color{red} red}), even though it is not present in the original problem. However, note that in this setting $\x_S \sim P$ is from a different and unknown distribution than $\x_T$ (as shown by \fref{fig:t0}); the input \emph{spaces} are however both of the same dimension (namely, $\{0,1\}^2$). }
\end{figure}

Consider \Fig{fig:t0} which illustrates as a probabilistic graphical model transfer-learning under the assumption that tasks $Y_S$ and $Y_T$ are intrinsically independent of each other (our assumption). We could conclude that the decision for $Y_T$ does not require or benefit from joint-modeling with $Y_S$; since
\begin{align*}
	\label{eq:ia_0}
	P(Y_T|\x_T) &= P(Y_T|\x_T,y_S) \\
	&= P(Y_T|\x_T) \cancel{\color{gray} P(\x_T)/P(\x_T)\sum_{y_S} P(y_S|\x_S)P(\x_S) P(\x_S)}  
\end{align*}
($Y_S$ is marginalized out, and observed $\x$-terms cancel out); i.e., it is clear that the predictive distribution $P(Y_T|\x)$ is not influenced in any way by the source task. 


The key observation is that at prediction time we are not dealing with true distributions $P$, but (as indicated in Eq.~\eqref{eq:e2}) approximations $\Phat$. Further, we do not approximate $P(X_S)$ or have any direct means to do so (observations $\x_S$ are never available). Rather, we map $\x_T$ into $\dX_S$ and then to $\yp_S$ deterministically using $f$ and $h_S$, respectively. This is shown in the deterministic graphical model of Fig.~\ref{fig:tl_novel_0}. Probabilistic inference can then be represented by \Fig{fig:t1bis} where conditioning on both observations renders the other connectivity irrelevant. 


The link $\Yp_S \rightarrow \Yp_S$ appears due to the fact that, usually, $\Phat \neq P$ and thus potentially 
\begin{equation}
	\label{eq:nia_0}
	\Phat(Y_T|\x_T) \neq \Phat(Y_T|\x_T,\yp_S)
\end{equation}

\begin{figure}[h!]
	\centering
	\subfigure[$S$ource Problem ($y_S \equiv$ logical \textsc{and} function)]{%
		\includegraphics[width=0.4\textwidth]{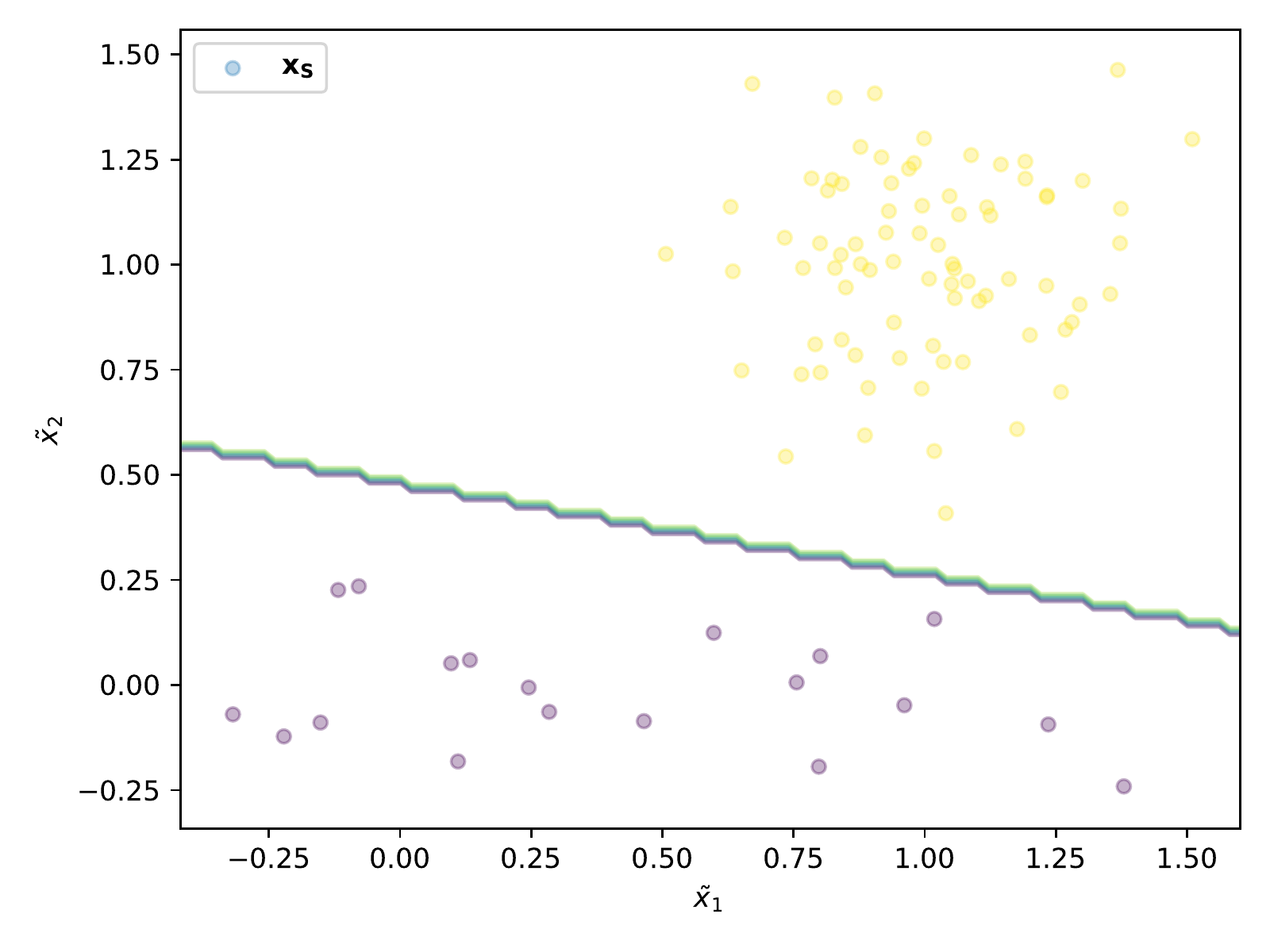}
		\label{fig:and}
	}\quad
	\subfigure[$T$arget Problem ($y_T \equiv $logical \textsc{xor} function)]{%
		\includegraphics[width=0.4\textwidth]{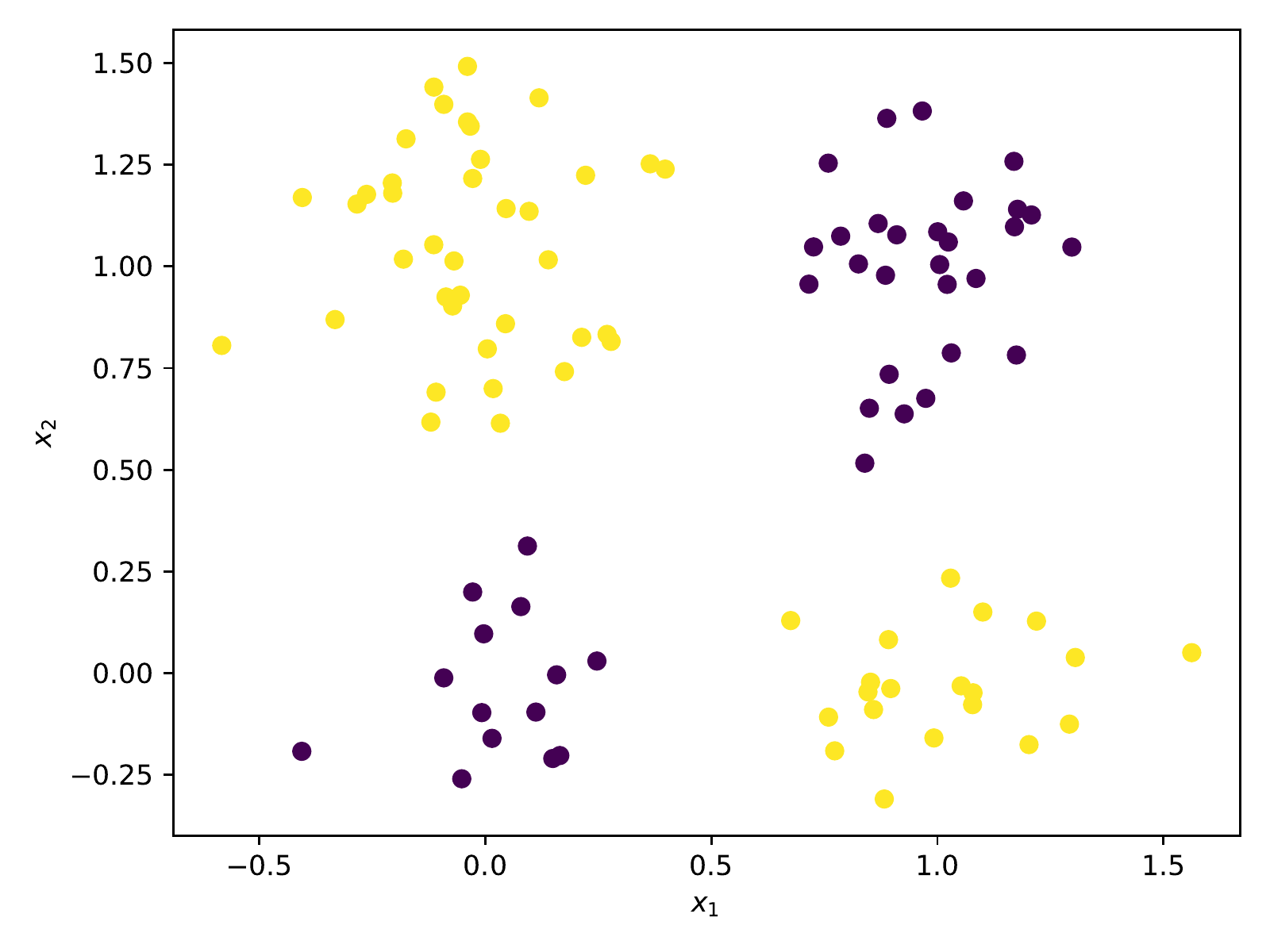}
		\label{fig:xor}
	}\quad
	\caption{\label{fig:D}Toy \fref{fig:xor} $T$arget and \fref{fig:and} $S$ource data sets; both generated independently of each other, as in Fig.~\ref{fig:modelB}, where both source domains are $\in \R^2$. Points in \fref{fig:and} (i.e., $\x_S \in \dX_S$; $\x_S \sim P_{\dX_S}$) have been faded to emphasise that we do not observe them; nor the decision boundary shown in that figure. We only have classifier/decision-rule $h_S : \dX_S \mapsto \dY_T$ (the decision rule). The goal is to map $\x_T$ (points in \fref{fig:xor}, i.e., $\in \dX_T$) into $\dX_S$ by building an appropriate mapping $f : \dX_T \rightarrow \dX_S$. This mapping can be considered a success if feature $\yp_S = h_S(f(\x_T))$ leads to increased performance for a fixed target model class $h_T : \dX_T \times \dY_S \rightarrow \dY_T$.}
\end{figure}

This tells us that we should take into account $\yp_S$ as there is clearly conditional dependence. This dependence has been manufactured artificially by a modeling failure. \Fig{fig:modelB} provides an an example of this mechanism in the transfer learning context (extending Fig.~\ref{tab:xor_1} from earlier, which was only a multi-label problem; this time $P(\rX_S)$ and $P(\rX_T)$ may have different distributions): predicting the \textsc{xor} label becomes conditionally dependent on the prediction of the \textsc{and} label, if minimal linear modeling is used; in order to obtain best results. This example is further extended in Fig.~\ref{fig:D}.

Naturally, this artificial dependence only arises when $h_T$ has insufficient capacity or otherwise is not sufficiently effective. 
To the contrary, then $\yp_S$ may be uninformative or otherwise have no influence on the decision of $h_T$. This may also happen if $h_S$ performs no useful function (e.g., outputting a constant or random noise), or otherwise fails to add any predictive power above what $h_T$ already offers. Although note that this may occur even in if the tasks were intrinsically \emph{dependent}, such as in low-noise scenarios when the target task can be solved easily and efficiently by $h_T$ alone. 


Hence the main assumption: that $h_S$ provides some useful predictive capacity for some [source] task. And thus the challenge: find $f$ such that $\yp_S = h_S(f(\x_T))$ is an effective feature to predicting $y_T$, \emph{in addition to} $\x_T$. We are particularly interested and likely to benefit from multi-label/multi-output problems where $\ypred_S$ is a feature/label vector, simply because we are able to obtain a larger number of features from a single problem. Thus, henceforth we generalize to this case where $\ypred_S \in \R^m$.

\subsection{Considerations for the mapping-function}
\label{sec:stuff}

In Transfer Chains (recall: \Eq{eq:e1}--\eqref{eq:e3}) we use a fixed and unobserved source model $h_S : \dX_S \rightarrow \dY_S$ already trained on a data to which we no longer have access and which we assume has no underlying measurable similarity to our target data set. We only know that $h_S$ takes $|\dX_S|$ inputs and $|\dY_S|$ outputs. We rely on a mapping function $f : \dX_T \rightarrow \dX_S$ to put $\dX_T$ into the right shape, and place it into a space that will provide a useful transformation. This is a crucial matter, since if we cannot make use of $h_S$ in any way, there can be no justification for our approach. 

According to the development so far, once removing the assumption of task similarity, the main utility left in $h_S$ is as its potential use as a non-linear function to improve overall predictive capacity. Existing deep neural networks approaches are an obvious choice, providing any amount on non-linear capacity, and training them is an extensively studied task. However, as motivated earlier, that is still an inherently data-centric view and ignores a vast wealth of knowledge and processing capacity stored in the form of innumerable existing models. 







Our point of departure is the minimal assumption that the source model provides a useful map for \emph{some} problem. Supposing this, we wish to put it to work for our target problem. The first obstacle is that both the distribution and dimensions of this source problem are already fixed.  

We might try to `disguise' our target input instance $\x_T$ as a source instance, suggesting the use of a transport map \citep{villani2009optimal} between the target and source distributions, $P(X_T) \mapsto P(X_S)$. Alas, we have neither those distributions nor the data with which we can form an approximation (at least, not $P(X_S)$). And even if we did, transport maps are an expensive undertaking that do not directly address our focus point which is the predictive performance of target model $h_T$. 

Therefore, we directly seek out the functional part of the $\dX_S$-space wrt decision-making, which is near the decision boundary in classification (in the case of regression models, and more generally -- the decision surface). 

Consider \Fig{fig:D}. 
The figure frames the problem in a visual way: how to map points $\x^{(i)}_T$ from the right (target) space into the left (source) space, in a way that $\ypred_S = h_S(f(\x_T))$ is a useful feature for predicting $\y_T$. 
In the context of Transfer Chains, the problem can be posed in another way: how to design map $f$ to maximize the performance $J$ of target model $h_T$ on test data (supposing a fixed target and source models $h$)? We can cast this as a generic optimization problem
\begin{equation}
	\label{eq:generic_opt}
	f^* = \argmax_{f \in \dF} J(f)
\end{equation}
with performance measure 
\begin{equation}
	\label{eq:J}
	J(f) \equiv J(\ypred_S,\y_T|\x_T) \appropto \Exp[\ell^{-1}(\rY_T,f)]\text{;}
\end{equation}
in other words, the importance of features $\ypred_S$ generated via map $f$ and source model $h_S$, wrt predicting the values $\y_T$ (alongside original inputs $\x_T$). 

Since $\ypred_S$ can be treated as a vector of features for the target model $h_T$, we can consider existing tools for assessing feature importance, e.g., filter metrics like mutual information 
as well as more computationally-intensive wrapper methods (involving the (re)-training of $h_T$). 
In our experiments we indeed experiment with such measures: namely mutual information, denoted $J_I(f) = I(\ypred_S,\y_T)$; and internal cross-validation returning an average loss according to loss function $\ell$ on predictions $\ypred_T$ vs $\y_T$, denoted as $J_\ell(f)$ (in experiments, we use $0/1$-loss (\Eq{eq:01}) but this choice is problem-dependent). Both are fairly standard filter- and wrapper-approaches (respectively) for feature selection. 
Essentially we are trying to maximize and measure how much `artificial' dependence has been manufactured during the training process. 


Since the main purpose of Transfer Chains is to leverage existing non-linearities in $h_S$, we restrict mapping function $f \in \dF$ to a simple linear transformation:  
\begin{align}
	\label{eq:model_class}
	\xtest_S &= f(\x_T) = \U^\top\x_T + \vu 
\end{align}


Even with this simplification, finding the best search mechanism for $f \in \dF$ is an extensive problem beyond the scope of this paper. In experiments we use vanilla hill-climbing methodologies; with proposal function adding Gaussian noise to $\U$ and a given budget of iterations (specified in the experiment section below). 

\begin{figure}[h!]
	\centering
	\subfigure[Linear map $f$ into $\dX_S$, points colored as $\yp_S^{(i)}$ (\textsf{AND})]{%
		\includegraphics[width=0.45\textwidth]{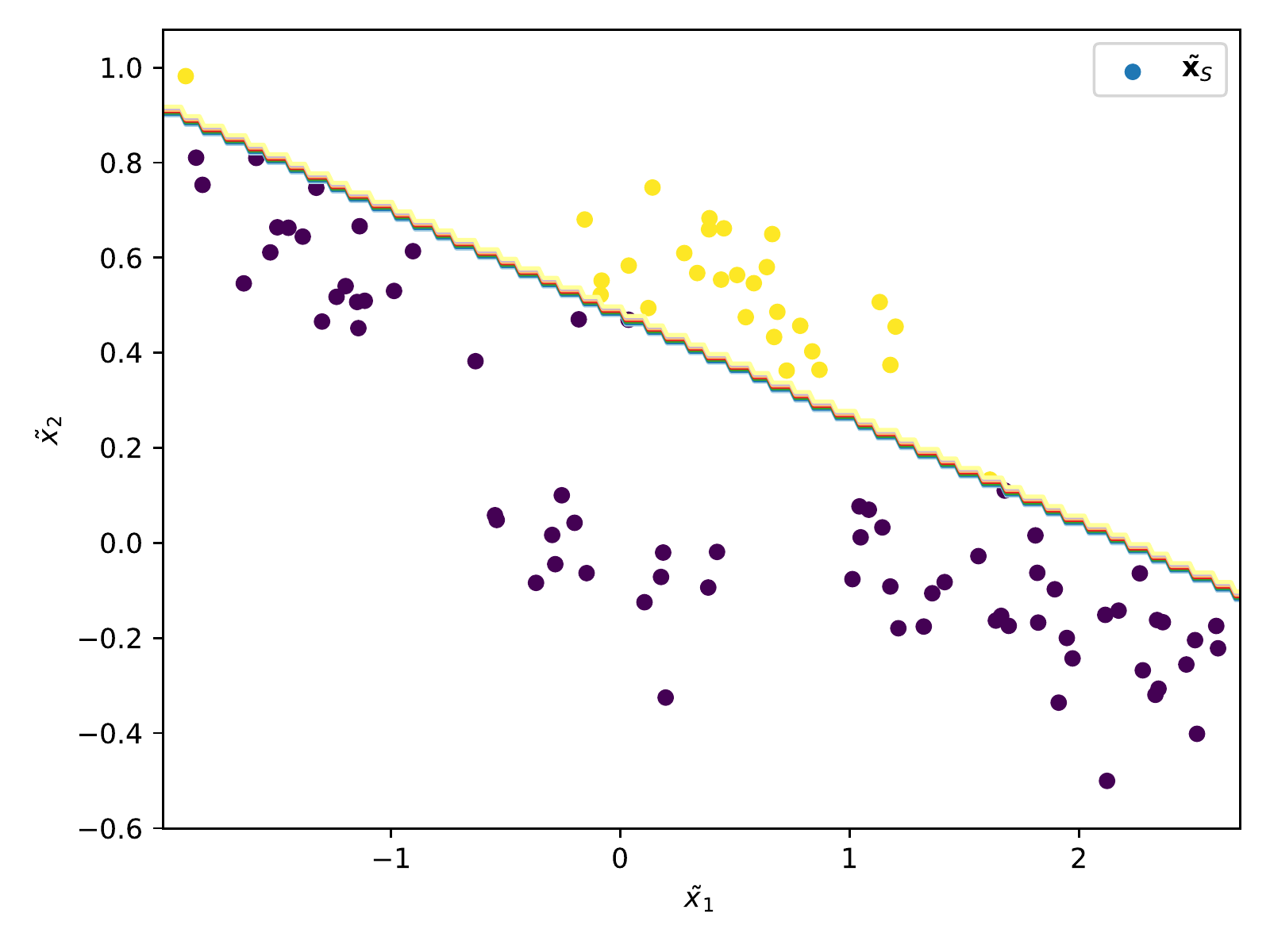}  
		\label{f:g}
	}
	\subfigure[Linear map $f$ into $\dX_S$, points colored as $\yp_T^{(i)}$ (\textsf{XOR})]{%
		\includegraphics[width=0.45\textwidth]{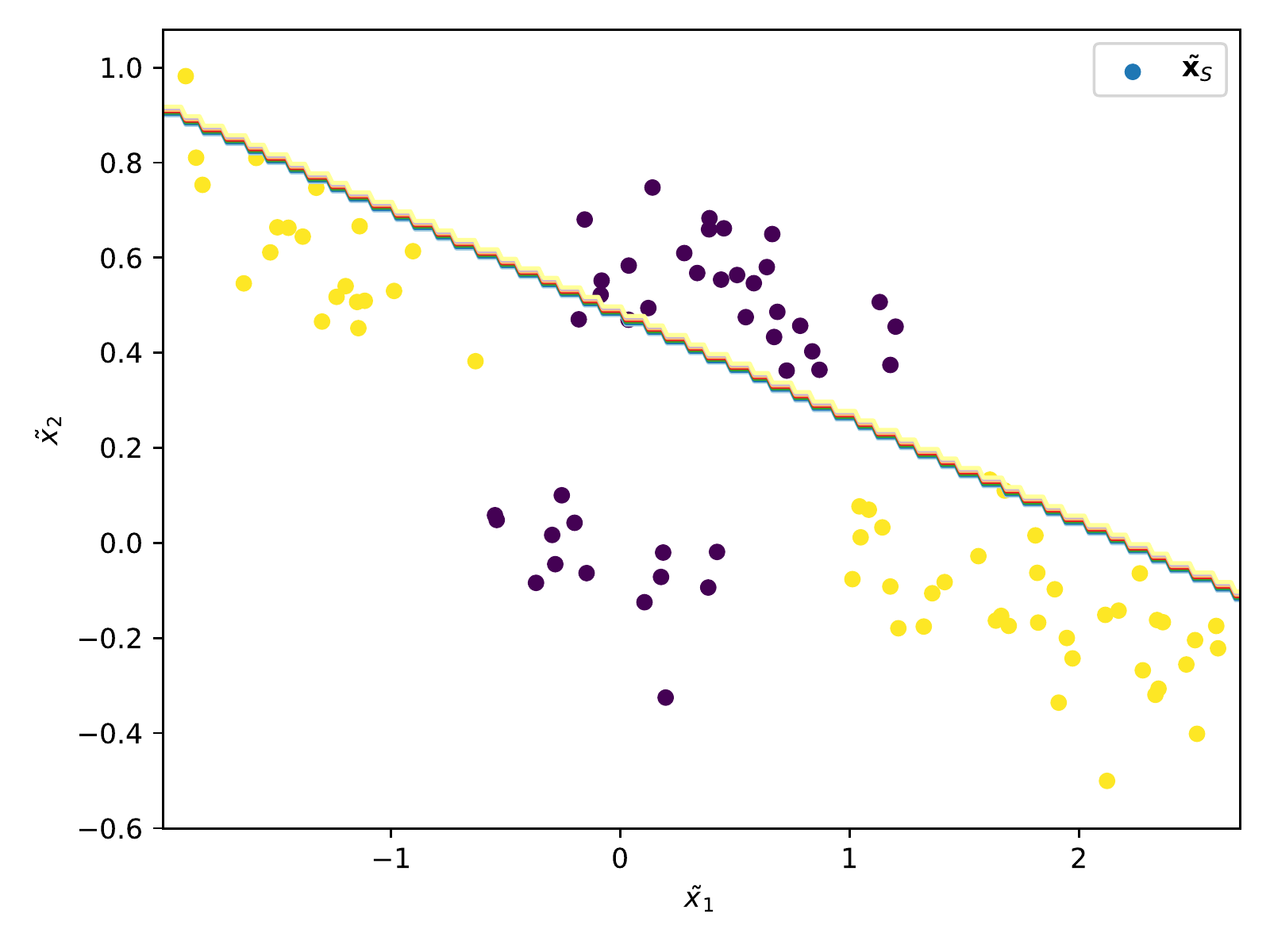} 
		\label{f:r}
	}
	\caption{\label{fig:projections}Proof of concept of Transfer Chains on target problem \textsf{XOR} using source problem \textsf{AND} (continuing from \Fig{fig:D}). Points $\xtest^{(i)}_S \in \dX_S$, obtained from $\xtest^{(i)}_S = f(\x^{(i)}_T)$, where $f \in \dF$ according to a simple hill-climbing search, from \Eq{eq:generic_opt} (via $f$ in \Eq{eq:model_class}) using performance measure $J_{0/1}(f)$ (exact match, i.e., --equivalently-- inverse $0/1$ loss). In \fref{f:g}, the color/label corresponds to $\yp^{(i)}_S = h_T(\xtest^{(i)})$. 
	In \fref{f:r}, the color/label corresponds to $y^{(i)}_{T}$. 
		}
\end{figure}

A high-level overview of the framework is given in \Code{code:algorithm}. 
\Fig{fig:projections} provides a demonstration and insight on a toy example. 

\begin{algorithm}[h!]
	\begin{algorithmic}[1]
		\Procedure{train}{$h_S$,$\{(\x^{(i)}_T,\y^{(i)}_T)\}_{i=1}^n$}
			\State $f \gets \argmax_{f \in \dF}J(f)$ \comment{Search; \Eq{eq:generic_opt}}
			\State $\xtest^{(i)}_S = f(\x^{(i)}_T)$ \quad $\forall i = 1,\ldots,n$ \comment{Mapping; \Eq{eq:model_class}, and \Eq{eq:e1}}
			\State $\ypred^{(i)}_S = h_S(\xtest^{(i)}_T)$ \quad $\forall i = 1,\ldots,n$ \comment{Feature creation; \Eq{eq:e2}} 
			\State Fit $h_T : \dX_T \times \dY_S \rightarrow \dY_T$ using dataset $\{[\x_T,\ypred_S]^{(i)},\y_T^{(i)}\}_{i=1}^n$
			\State \Return $f,h_T$
		\EndProcedure
		\medskip
		\Procedure{predict}{$\x_T,h_S,f,h_T$}
			\State $\xtest_S = f(\x_T)$
			\State $\ypred_S = h_S(\xtest_S)$
			\State \Return$\ypred_T = h_T(\x_T,\ypred_S)$ \comment{\Eq{eq:e3}}
		\EndProcedure
	\end{algorithmic}
	\caption{\label{code:algorithm} Transfer Chain: A high-level overview}
\end{algorithm}

\subsection{Experimentation}
\label{sec:experiments}


We conduct experiments to investigate if Transfer Chains can offer promise in practice for model-agnostic cross-domain transfer learning. As target problems we use benchmark multi-label classification datasets, listed in \Tab{tab:datasets} alongside their corresponding source problems (drawn from the same pool of benchmark sets) It is important to recall that only black-box source models are used (not the source data) for the target problems. To that end, we train multi-label random-forest classifiers on source problems and, without loss of generality, pool them together as a single source model $h_S$. For example, the source model we have available when using \textsf{Music} as a target data set, is as  
\[
	\ypred_S =  h_S(f(\x_T)) = [h_\textsf{Scene},h_\textsf{Yeast}](\xtest_S)   
	= [\yp_{1,1},\ldots,\yp_{1,6},\yp_{2,1},\ldots,\yp_{2,14}]
\]
We do not use \textsf{Birds} or \textsf{Music} in the same experiment as each other's source or target because their domains are related; namely both are audio. Even though there is no other similarity beyond the audio domain, we wish to test in the context of cross-domain transfer learning in the strict sense. Multi-dimensional source datasets are important to our approach, in order to broaden the information flow towards target models (more source outputs implies more features for those models). 

\begin{table}[h!]
	\centering
	\caption{\label{tab:datasets}Multi-label data sets ($d$ attributes and $m$ binary labels) and their corresponding source problems used in experiments (as black-box models $h_S$). Available from \url{https://www.uco.es/kdis/mllresources/}.}
	\begin{tabular}{lllll}
		\toprule
			Target Dataset    & $n$  & $d$  & $m$ & Domain  \\
		\midrule                     
			\textsf{XOR}      & 200    & 2    & 1   & Logical/Synthetic \\
			\textsf{AND}      & {\color{gray}100}    & 2    & 1   & Logical/Synthetic \\
		\midrule                     
			\textsf{Music} & 593  & 72  & 6   & Audio/Music \\
			\textsf{Scene}    & 2407 & 294 & 6   & Image/Scene \\
			\textsf{Birds}    & 645  & 260  & 19  & Audio/Birdsong \\
			\textsf{Yeast}    & 2417 & 103 & 14  & Microbiology \\
		\bottomrule
	\end{tabular}
	\begin{tabular}{l}
		\toprule
			Source Problem \\
		\midrule
			\textsf{AND} \\
			\\
		\midrule
			\textsf{Scene}, \textsf{Yeast} \\
			\textsf{Music}, \textsf{Yeast} \\
			\textsf{Scene}, \textsf{Yeast} \\
			\textsf{Music}, \textsf{Scene}, \textsf{Birds} \\
		\bottomrule
	\end{tabular}
\end{table}

Transfer learning should offer some advantage vs standalone fully data-driven target models $h_T$. In particular, we will look for: best predictive performance, best initial predictive performance, and fastest increase in predictive performance (mentioned by \cite{TransferLearningBook} and others as three desiderata for transfer learning). Target model classes $h_T$ used in the experiments are described as follows where we use the concept of \emph{step} to be able to compare at different points of training:



\begin{itemize}
	\item \textsf{SLP}: \textbf{Single-Layer Perceptron}, one logistic regression model per output trained via SGD \\
		At each step: An additional 100 iterations of SGD for each base learner. 
	\item \textsf{ECC}: \textbf{Ensemble of Classifier Chains}, each chain in a random order\\  
		At each step: A new base model (random chain) is added; 
	\item \textsf{MLP}$_v$: \textbf{Multi-Layer Perceptron} (architectures $v$ denoted below) \\
		At each step: An additional $100$ iterations of SGD.
	\item \textsf{RLP}$_v$: \textbf{Random Layer Projection}: Initialized as \textsf{MLP}$_v$ but \emph{not trained} \\ 
		At each step: trial new random weights; keep them if performing better (hill climbing on $J_\ell$); 
	\item \textsf{TC}$_v$: \textbf{Transfer Chain}. Subscripts and layers as per \textsf{MLP}. 100 initial search iterations $f \in \dF$  \\
		An additional 100 iterations of SGD for base learners 
	\item \textsf{ETC}: \textbf{Ensemble of Transfer Chains}, Ensemble of random \textsf{TC} \\
		At each step: A new transfer chain (random $f \in \dF$; then 100 iterations SGD) is added.
\end{itemize}
Where $v$ determines the architecture, such that  
\begin{itemize}
	\item $v=2$: 2 hidden layers of 100 units each, 
	\item $v=1$: 1 hidden layer of 400 units, and 
	\item $v=0$: 0 hidden layers (thus \textsf{SLP} $\equiv$ \textsf{MLP$_0$}).
\end{itemize}
As model-agnostic methods, we can consider any base model. We select stochastic gradient descent (SGD) since it provides a clear view of running time divided up into iterations, and thus provide a clearer picture of accuracy vs computational time invested. We use a weight decay penalty $\lambda=0.05$ throughout. For all ensemble learners a new model (rather than additional iterations) is added at each step.

\begin{figure}[h!]
	\centering
	\includegraphics[width=0.4\textwidth]{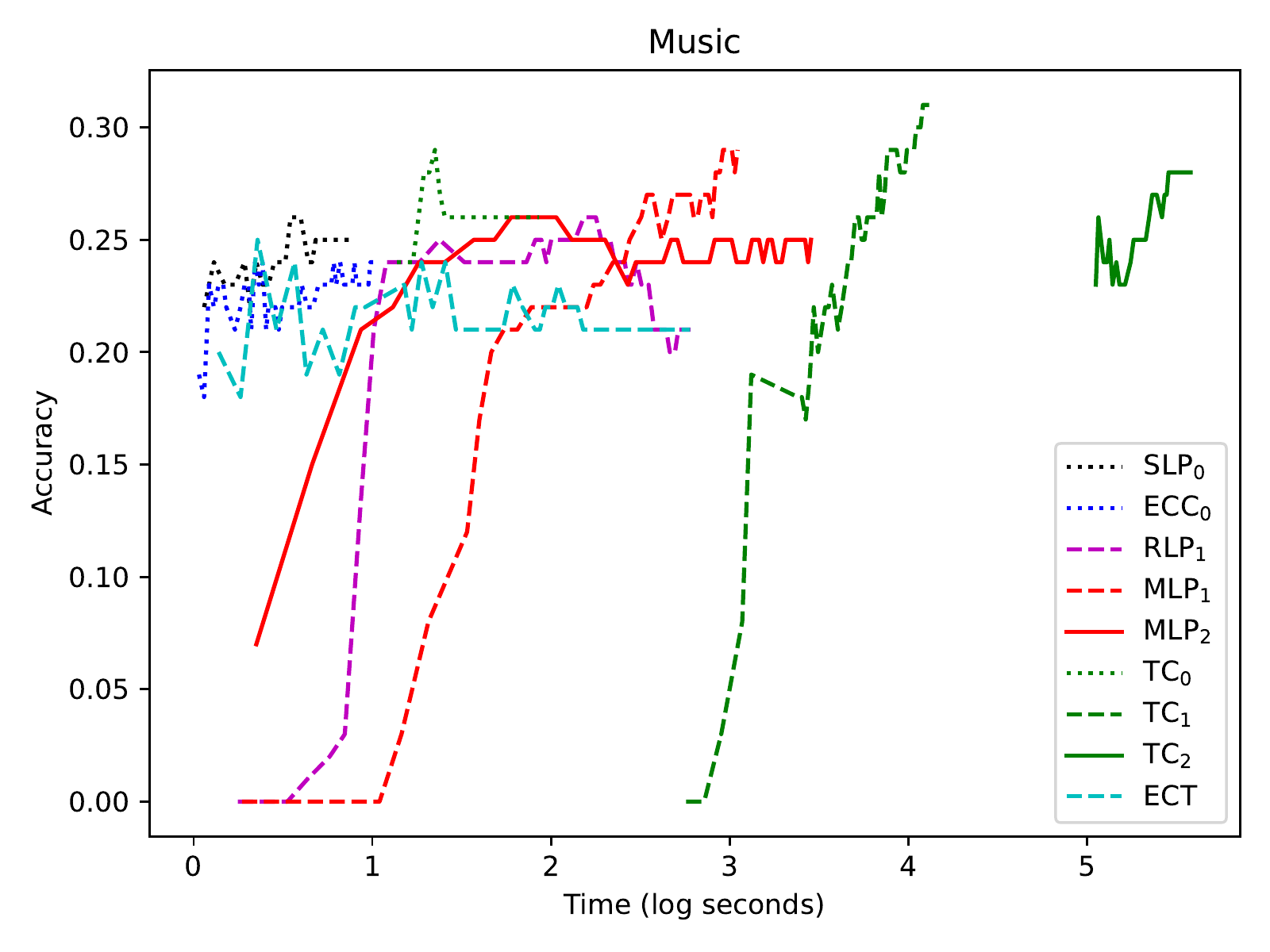}
	\includegraphics[width=0.4\textwidth]{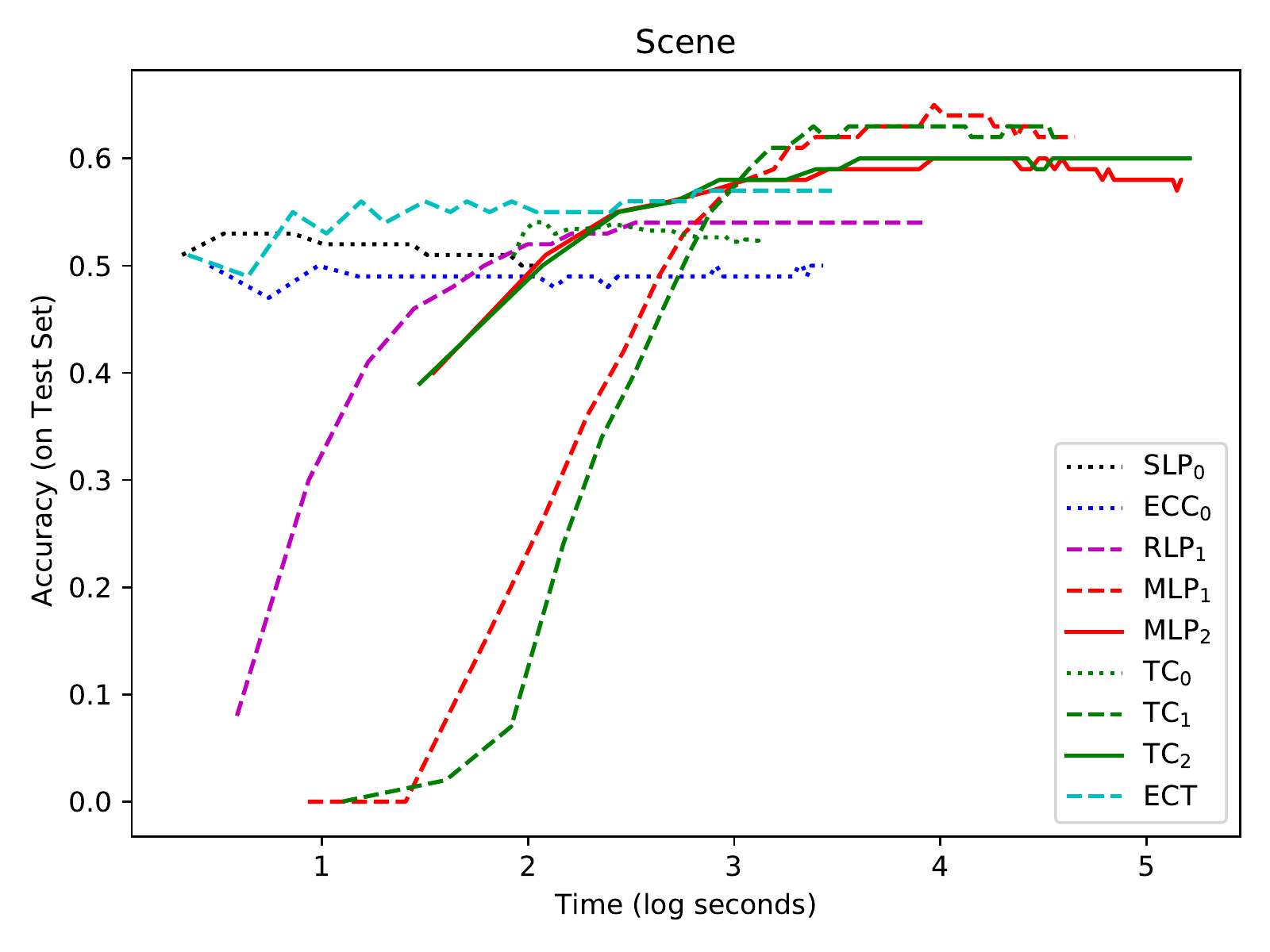}  \\
	\includegraphics[width=0.4\textwidth]{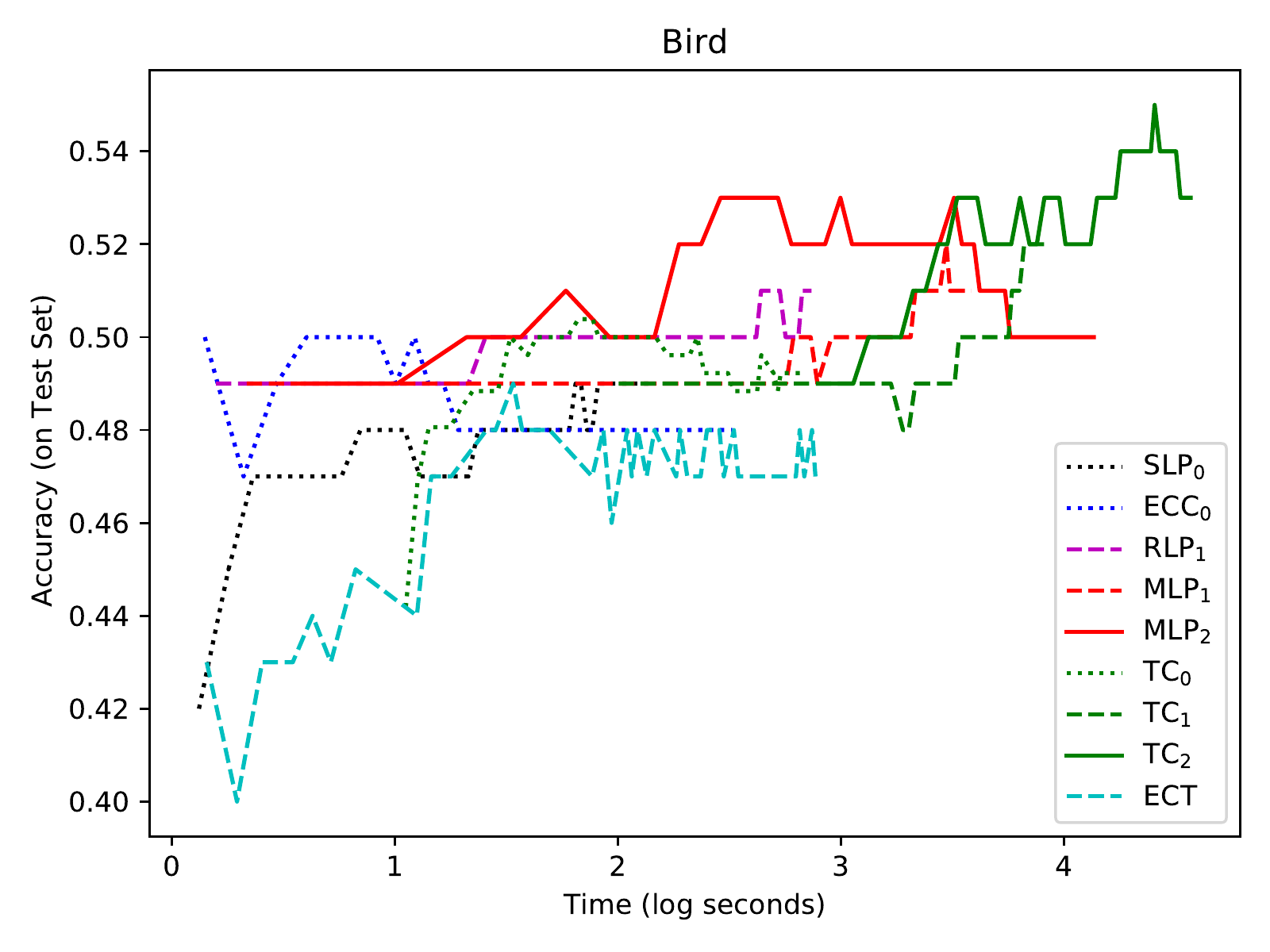} 
	\includegraphics[width=0.4\textwidth]{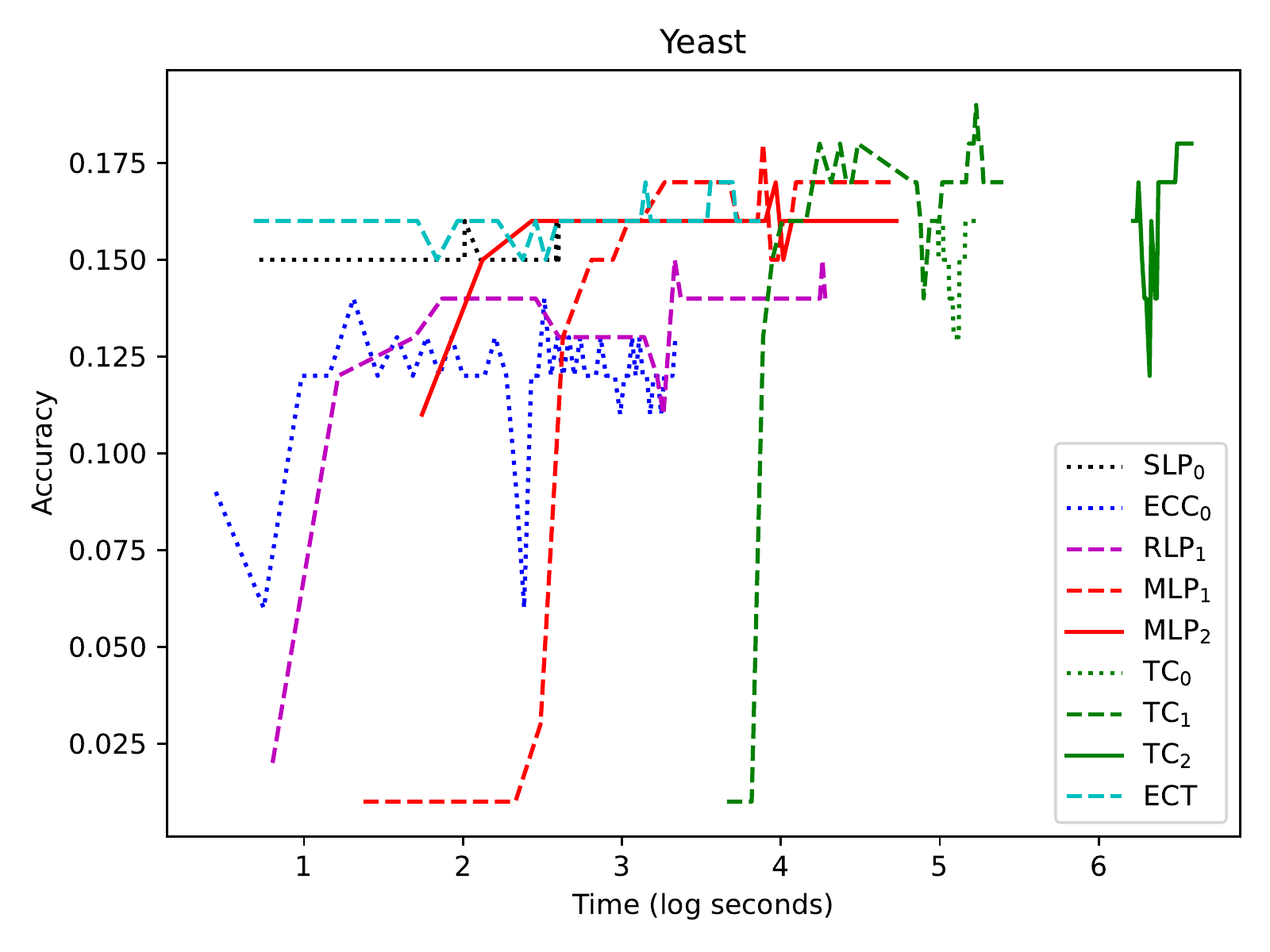}
	\caption{\label{fig:exp4full}The performance of target models on multi-label datasets with a shuffled 60:40 train-test split, over 50 steps. Subset accuracy/exact match is given on the test set versus training time.}
\end{figure}

Experimental methodology: we run each method over 50 steps on the datasets (listed in \Tab{tab:datasets}), recording accuracy (exact match; inverse $0/1$ loss) and computational expenditure at each step. Results are given in \Fig{fig:exp4full}. 

All methods are implemented in Python 
making use of the scikit-learn framework \citep{ScikitLearn} for base models (SGD and classifier chains) where default parameters are as such unless otherwise specified above. All experiments carried out on a laptop with Intel 1.80GHz processors. 


We do not compare to state-of-the-art methods in deep transfer learning which requires model introspection and manipulation and is thus not compatible with our assumptions and experimental setup. The main intention of these experiments is to provide a proof of concept supporting the discussion and development we put forward in this article, studying in particular the question of model-agnostic cross-domain transfer learning in a strict sense, and its link to multi-label learning.  




Results are given in \Fig{fig:exp4full}. We see that one of variety of Transfer Chains (either \textsf{ECT}, or \textsf{TC$_v$}) performs best on all datasets (equal best in the case of \textsf{Scene} in terms of best performance). There are a large number of effects to decode, from overfitting to capacity to the influence of the underlying learning algorithm (SGD, in all cases) and the source model. Certainly results are not overall better on all datasets in all settings; yet we can come back to the main message: taking a model-based approach, even though they are black-box models, rather than a purely-data driven approach, holds promise -- and this is a very interesting concept adding weight to the idea that a shift towards model-driven modeling rather than a purely data-based approach, can be an interesting opportunity to explore further. Further detailed discussion is given in the following section. 


\section{Discussion, Insights, and Implications}
\label{sec:discussion}


There is an important tradeoff between training the base models (i.e., the parameters of target model $h \in \dH$), and finding a good mapping function (i.e., $f \in \dF$), alongside the question of how many models (ensemble, or not) versus how much training time to invest in each of those models. These considerations are highly interdependent, since the effectiveness of the mapping function depends on the effectivness of the target model and vice versa. 

It is evident in Fig.~\ref{fig:exp4full}, as is also well known in the literature, that simple hill-climbing has a limited over a small number of iterations on a high number of dimensions. At the same time, although the effect is limited it may nevertheless be significant. 

Recall that our main aim is not to provide a new state-of-the-art method for transfer learning, because such methods assume task similarity, available source training data, or transparency of the source model. Our approach assumes none of these. It is already clear that task similarity is an important ingredient to success of transfer learning, just as label dependence is a major factor in the success of many multi-label and multi-task methods. However we have sought (in both cases) to highlight and demonstrate that it is not the \emph{only} ingredient and to develop a study around the question of other aspects. 

Therefore, although one can easily argue that transfer chains does not constitute an immediately attractive option in consistent efficient performance, its performance -- alongside the development and ablation studies earlier -- nevertheless raises interesting points of discussion and here we reflect upon these and the implications they can have in future research. 

\subsection{The potential of model-agnostic cross-domain transfer chains}

Our experimental results of \Sec{sec:experiments} support the hypothesis derived from our analysis (in \Sec{sec:lessons}): in the absence of task similarity, the gain for considering tasks together is arguably slight in some circumstances, yet non-negligible and even important in other cases. 

One argument can be that gains could be met by more careful fine tuning and hyper-parametrization of existing data-driven approaches, but there is another argument that we must take a step to the side in order to take a step forward: massive data-driven learning approaches are not tenable to those who do not have the data, and computational expensive to those who do. And even in the context of enormous reserves of data and computational resources to process it, it unrealistic to rely on this approach, and it is certainly not the case in human learning or in nature. 

From our developments, it appears apparent that much of the gain from the chain transfer is an addition of structural capacity and predictive power from the base model. Multi-layer networks with non-linear activation functions (represented by MLP) can provide enormous model capacity. However, providing some of this capacity via model-agnostic transfer appears to convey certain advantages. It some cases this is sufficient to reduce the need for (or at least depth of) back-propagation. 



If predictive architecture non-linear capacity is an important factor, it is true that random projections are cheap method to obtain this, and are well known to the machine learning literature in many forms. However, although their inner layers are produced instantly as random weight matrices (and, because of this) they are notoriously wasteful, and indeed in Fig.~\ref{fig:exp4full} exhibited limited success in the form of \textsf{RLP}.  


There is a qualitative difference in Transfer Chains, which uses random models only in the sense of selected from a pool of source models that are not likely to be related to the target task more than any other given task; the models themselves are not randomly initialized. Rather, since we make the assumption they were constructed by some machine learning framework we know their structure is efficient (formally: maximizing likelihood with some form of regularization) for at least some task and that they are already available. Hence also, we by this they are not wasteful; no additional creation or additional memory costs. The only challenge is having to find a relatively useful mapping to and from such models. 

We showed that this process is similar to artificially creating task dependence. This also revives the question: what is similarity between two tasks. Drawing two tasks randomly from the pool of all available tasks much have some underlying structural connection, even if it is just by accident. Naturally overfitting is then a risk (as with random projections and basis function expansion and so on) but with sufficient regularization (another inherent benefit from multi-task learning), a satisfactory fit can be obtained, thus recycling capacity, rather than learning it from scratch.

\subsection{Implications for learning from data streams with abrupt shifts in concept}
\label{sec:insights_streams}

Earlier (\Sec{sec:types}) we mentioned the case of data streams; a well-known area of research building methods to deal with data arriving constantly in a theoretically-infinite stream. Within this area, adapting to concept drift is the widely-known and relevant challenge -- essentially a special case of transfer learning -- of identifying and jettisoning knowledge which is no longer relevant and learning the new concept as quickly as possible (at the same time avoiding catastrophic forgetting, i.e., of failing to retain knowledge which will be useful again in the future (a reoccurring concept)). 

While a full consideration is beyond the scope, we can concisely say (references provided in \Sec{sec:types}) that the state of the art methods in the data-stream literature will typically look to detect drift, destroy existing structure which represents knowledge from the previous concept in order to free up memory, and allow it to regrow/reform based on the new concept -- i.e., to actively promote forgetting. Following this argument, when drift is abrupt and complete (old concept bears no similarity to current concept), one could justify abandoning the existing models completely. 

Our study indicates that such a reaction (destroying models referring to a no-longer valid concept) would be a major mistake even in cases of total and abrupt drift. Rather, one should aim to incorporate the old models into the current decision process for the new concept. This is precisely the best opportunity for methods such as Transfer Chains: following the start of a new concept, there is a small amount of target data, meaning that both other mechanisms we looked at -- the `James-Stein effect' and the `chaining effect' are likely to render most benefit, and further more the presence of an model is already available and in place from the previous concept. 

Furthermore, using a model in this way would be immediately relevant in the case of a recurrent concept. Recurrent concepts are already studied, but what our results indicates in that using models can be reused even if not explicitly appearing again in the stream in the same form.  
Yet we could suspect this mechanism could already be in play in data streams. Concept drift detection is notoriously imprecise, and so models are kept around much longer than they need to be. Further analysis in the context of data streams would be needed to properly develop this hypothesis; even if such a study is outside the scope of this article. 


%

\subsection{A fresh look at connections to human and biological transfer}
\label{sec:insights_neuro}

There is no novelty in drawing connections to human learning when we discuss transfer learning, as an immense amount of literature exists (much of it since before machine transfer learning became known or popular). But we can highlight that the kind of transfer learning in humans is, like the approach we investigate, also mainly model-driven learning. 
Indeed, it is now well known that the human brain is not a data storage device in the sense of storing original data for faithful recall later \citep{HowTheMindWorks}, at least not in the same sense as a database or csv file. On observation and interaction our brains develop neural pathways for processing information, and these pathways can be developed regardless of objective task similarity. For example, learning a foreign language may benefit achievement in mathematics \citep{stewart2005foreign}, and the study of mathematics is commonly motivated beyond contexts where it can be directly applied. 
This raises the same question we have studied: there is a non-trivial overlap between the concepts of transfer based on knowledge-based similarity, or of structural-similarity, or simply a transfer of capacity. 




Indeed, one view is that chain transfer essentially repurposes a source model, and gain an effect for which that model was not initially designed (again: transfer -- or simply making use of -- existing capacity). We can find this throughout biology, and it is clearly observed in the Covid19 pandemic: global behavior was modified with planet-wide impact via a relatively simple mechanism (a virus). Admittedly, the ensuing behaviour was not a target model of the virus, and much alteration of global activity was specifically against its own payoff metric (of reproduction), but it provides ample demonstration of the scale of leverage on/repurposing of a complex pre-existing structure by a comparatively simple vector. 


\subsection{Limitations versus future potential: a place for model-agnostic cross-domain transfer}

With the development and study of transfer chains we pushed towards the development of more model-driven learning approaches, moving a way from purely data-driven learning. Such an approach provides a novel methodology towards model re-use and recycling, and away from single-use models involving memory-wasteful and computationally-expensive from-scratch techniques. It is worth restating the main benefit: transfer learning that is employable using off-the-shelf black-box modules, not necessarily of a particular model class or framework.  




We have also acknowledged that explicitly denying aspects of task similarity comes potentially at a major cost in terms of accuracy during transfer learning. As a short-term goal it may not be the most practical way to achieve improvement in a target model. 
However, under a larger and more long-term view, its potential is greater, and one could envision a large library of source models to select from.  Due to the model-agnostic and model-driven nature, this could foster more easily collaboration among practitioners without compromising privacy of data sources. 



There are many avenues for improvement. We have so far only used relatively simplistic off-the-shelf base learners and hill-climbing methodologies.  
We can speculate that increasing the complexity of these methodologies as well as increasing the number of source tasks have improve the accuracy on the target task via a transfer chain, but further research would be needed to confirm this. 


We have only looked at multi-label classification for the source and target tasks. Having multiple outputs is useful in overcoming an information bottleneck, but there is a vast array of task that could be considered: multi- and structured-output regression, time series, data streams, reinforcement learning, etc. In particular, we did not yet consider recurrent architectures, but these architectures are still used in practice with learning regimes other than those based on gradient descent, including an area of research known as reservoir computing \citep{ReservoirComputing}. This would be a clear candidate for future investigation, alongside reinforcement learning. In transfer chains, not only does the source data not need to be similar to the target data, but the task definition does not need to be the same either; such that recurrent connections among decision tree-regression models could be used as part of the representation for reinforcement learning. Future work will investigate such possibilities.  



\section{Conclusion}
\label{sec:conclusion}

In this article we studied, developed, and leveraged mechanisms in multi-label and transfer learning other than task similarity, which is usually seen as a fundamental motivating assumption in these areas; removing it was a central point of our study. Namely, we looked initially in the multi-label case (which involves a large corpus of literature), where task similarity takes the form of label dependence. By removing the assumption of label dependence, we brought to surface theoretical insights on other mechanisms, such as effects of regularization and additional capacity that arise naturally in connecting different tasks. And we provided an empirical study to isolate and examine these effects in practice in benchmark multi-label datasets. We showed that these mechanisms, which have been hitherto underappreciated or even disregarded completely, can significantly impact the performance in multiple task systems. We made use of these findings to develop an approach to the extremely challenging goal of model-agnostic (black-box) model-driven (no source data) cross-domain (no task similarity) transfer learning. Despite this difficult setting, with our methodology of \emph{transfer chains} inspired from related concepts in the multi-label literature, that arose from our study, we were able to demonstrate attractive results on a number of standard data sets, in most cases out-competing the established methods. 

We did not question the important positive association between task similarity and predictive accuracy in transfer learning. Thus, by removing the assumption of similarity, our goal was not to produce a new state-of-the-art in multi-label or transfer learning, but rather we provided greater understanding on the significance and potential of effects other than task similarity; we developed some theoretical insights behind these effects, and demonstrated and isolated them in practice. This is a relatively small step, but we argue it has important implications for future development. We discussed a few of these implications; in particular, we elaborated a discussion on adapting to concept drift in data streams, among other possibilities. 

We identified many areas for promising future developments of model-agnostic transfer learning, including using recurrent modular structures and reinforcement learning. 

\bibliography{../../multilabel,../../datastreams,../../my_publications,../../erc.bib}

\end{document}